\pdfoutput=1
\documentclass{article}

\PassOptionsToPackage{numbers, compress}{natbib}
\usepackage[preprint]{neurips_data_2024}
\usepackage[utf8]{inputenc} % allow utf-8 input
\usepackage[T1]{fontenc}    % use 8-bit T1 fonts
\usepackage{hyperref}       % hyperlinks
\usepackage{url}            % simple URL typesetting
\usepackage{booktabs}       % professional-quality tables
\usepackage{amsfonts}       % blackboard math symbols
\usepackage{nicefrac}       % compact symbols for 1/2, etc.
\usepackage{microtype}      % microtypography
\usepackage{xcolor}         % colors

\usepackage{caption}
\usepackage{graphicx}
\usepackage{enumitem} % adjust itemize left margin
\usepackage{xspace}
\usepackage{amsmath}
\usepackage{amssymb}
\usepackage{multirow}
\usepackage{algorithm}
\usepackage{wrapfig}
\usepackage{subcaption}
\usepackage{bm}
\usepackage{listings}
\usepackage{xcolor}
\usepackage{lipsum}

\usepackage{colortbl}
\usepackage{pifont}
\usepackage{authblk}

\usepackage{amssymb} % For checkmark symbol
\usepackage{graphicx} % For scaling the table to fit page width
\usepackage{booktabs} % For better spacing in the table
\usepackage{colortbl} % For coloring tables
\definecolor{lightblue}{rgb}{0.68, 0.85, 0.9} % This creates a light blue color
\usepackage{array}    % For table column formatting
\usepackage{pifont}
\usepackage[export]{adjustbox}
\usepackage{algpseudocode}
\usepackage{tcolorbox}

\usepackage{color}

\definecolor{softyellow}{rgb}{0.98, 0.98, 0.82} % Soft yellow
\definecolor{tbgray}{gray}{.92}

\usepackage{longtable}
\usepackage{stackengine}

\stackMath

\title{MMAU: A Holistic Benchmark of Agent \\ Capabilities Across Diverse Domains}

\author{Guoli Yin\thanks{Equal contribution.} \hspace{0.3cm} Haoping Bai* \hspace{0.3cm} Shuang Ma* \\
        Feng Nan\hspace{0.3cm} Yanchao Sun\hspace{0.3cm} Zhaoyang Xu\hspace{0.3cm} Shen Ma\hspace{0.3cm} Jiarui Lu\hspace{0.3cm} Xiang Kong\hspace{0.3cm} Aonan Zhang\hspace{0.3cm} \\ Dian Ang Yap\hspace{0.3cm}
        Yizhe Zhang\hspace{0.15cm} Karsten Ahnert\hspace{0.15cm} Vik Kamath\hspace{0.15cm} Mathias Berglund\hspace{0.15cm} Dominic Walsh\hspace{0.15cm} Tobias Gindele\hspace{0.15cm} Juergen Wiest\hspace{0.15cm} Zhengfeng Lai\hspace{0.15cm} George Horrell\hspace{0.15cm}
        Xiaoming Wang\hspace{0.3cm} Jiulong Shan\hspace{0.3cm} Meng Cao\thanks{Senior authors.}\hspace{0.3cm} Ruoming Pang$^{\dagger}$\hspace{0.3cm} Zirui Wang$^{\dagger}$} 
\affil{Apple Inc.}

\begin{document}

\maketitle

\vspace{-1cm}
\begin{figure}[ht]
    \centering
    \includegraphics[width=\linewidth]{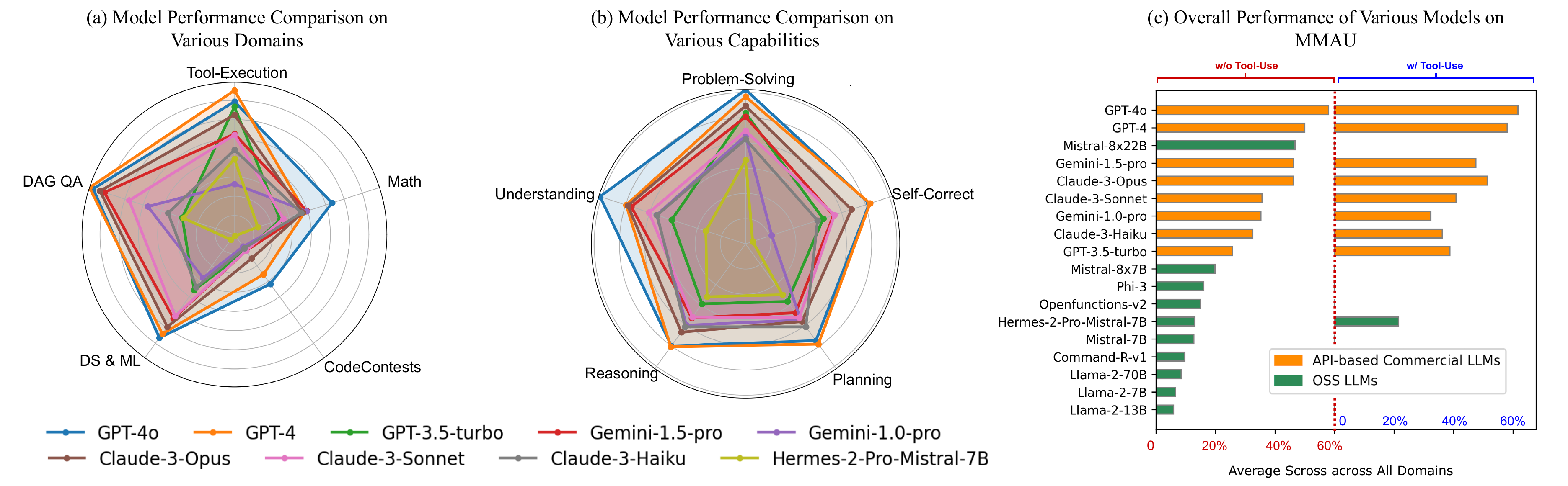}
    \caption{Evaluation results across different models on MMAU. For clarity, this figure includes only a selection of representative models. The domain-centric, capability-centric, and overall evaluation results are aggregated from all 20 tasks in MMAU. For detailed per-task evaluations, please refer to Appendix~\ref{sec:appendix_eval_results}.}
    % \vspace{-1.5em}
    \label{fig:mmau_combined_eval_charts}
\end{figure}

\begin{abstract} \label{sec:abstract}
Recent advances in large language models (LLMs) have increased the demand for comprehensive benchmarks to evaluate their capabilities as human-like agents.
Existing benchmarks, while useful, often focus on specific application scenarios, emphasizing task completion but failing to dissect the underlying skills that drive these outcomes. This lack of granularity makes it difficult to deeply discern where failures stem from. 
Additionally, setting up these environments requires considerable effort, and issues of unreliability and reproducibility sometimes arise, especially in interactive tasks.
To address these limitations, we introduce the Massive Multitask Agent Understanding (MMAU) benchmark, featuring comprehensive offline tasks that eliminate the need for complex environment setups. It evaluates models across five domains, including \textcolor{teal}{Tool-use}, \textcolor{teal}{Directed Acyclic Graph (DAG) QA}, \textcolor{teal}{Data Science and Machine Learning coding}, \textcolor{teal}{Contest-level programming} and \textcolor{teal}{Mathematics}, and covers five essential capabilities: \textcolor{orange}{Understanding}, \textcolor{orange}{Reasoning}, \textcolor{orange}{Planning}, \textcolor{orange}{Problem-solving}, and \textcolor{orange}{Self-correction}.
With a total of 20 meticulously designed tasks encompassing over 3K distinct prompts, MMAU provides a comprehensive framework for evaluating the strengths and limitations of LLM agents.
By testing 18 representative models on MMAU, we provide deep and insightful analyses.
Ultimately, MMAU not only sheds light on the capabilities and limitations of LLM agents but also enhances the interpretability of their performance. Datasets and evaluation scripts of MMAU are released at \url{https://github.com/apple/axlearn/blob/main/docs/research/mmau}.

\end{abstract}
\section{Introduction}\label{sec:intro}

\begin{figure}[t]
    \centering
    \includegraphics[width=\linewidth]{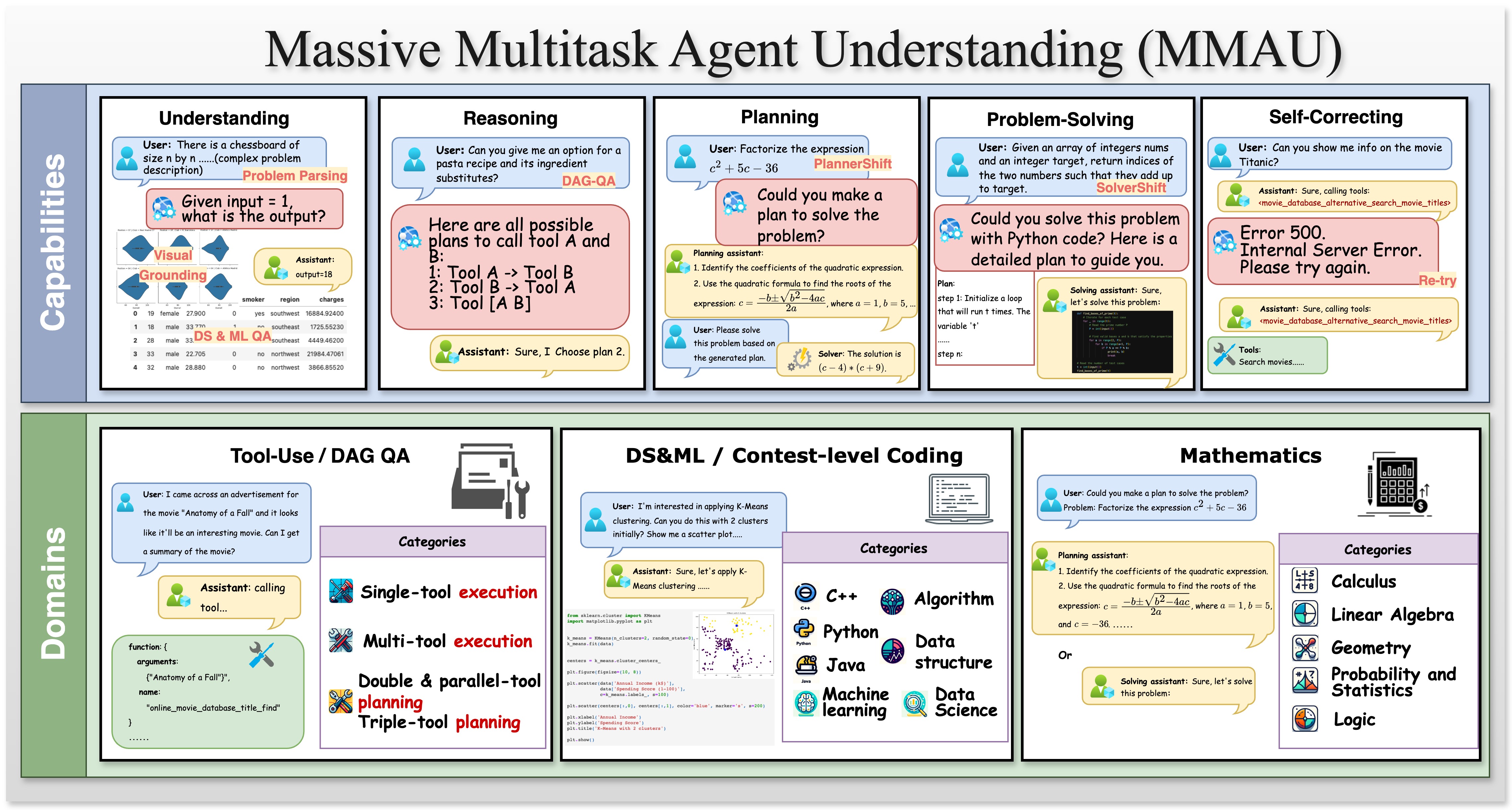}
    \caption{Overview of MMAU. MMAU is designed to provide both capability-centric evaluation (top) and domain-centric evaluation (bottom). It includes over 3K distinct prompts spanning 64 subjects and 5 domains. To evaluate the fundamental capabilities of LLM agents in a disentangled manner, we carefully designed 20 tasks aimed at decomposing these capabilities and assessing performance. Note: For clear visualization, the data examples and prompts here are simplified to illustrate an intuitive example. For the exact data examples and prompts, please refer to the Appendix~\ref{sec:appendix_data_examples}~\ref{sec:appendix_task_prompts}. }
    \label{fig:main-fig}
    % \vspace{-1.5em}
\end{figure}

Recent advancements in the field of AI have been marked by significant progresses in the development of LLMs. Particularly, one promising direction along this evolution is the ability of LLMs to perform as human-like agents~\cite{Coala:sumers2023cognitive}, i.e., understand complex contexts, reason and plan with complicated logic~\cite{tot:yao2024tree, cot:wei2022chain, llm+p:liu2023llm+, coh:liu2023chain}, make decisions, and utilize tools effectively~\cite{react:liu2023llm+, shinn2023reflexion, shen2024hugginggpt}. Consequently, the need for comprehensive benchmarks that evaluate LLMs as intelligent agents has become more and more important. 
% Such benchmarks are crucial for assessing the agent-like functionality of LLMs, ensuring they meet the demands of real-world applications. 

\begin{figure}
    \centering
    % \vspace{-1em}
    \includegraphics[width=.85\textwidth]{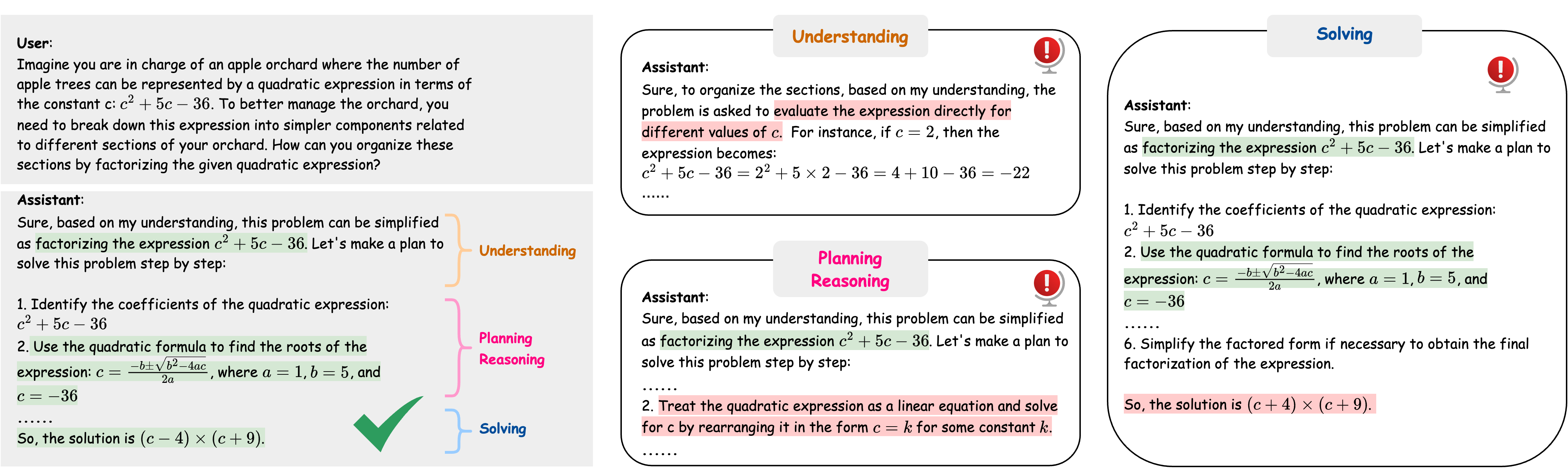}
    \caption{Different error types on a math problem.}
    % \vspace{-1em}
    \label{fig:error-type-example}
\end{figure}

% While there are existing benchmarks~\cite{mmlu:hendrycks2020measuring, yue2023mmmu, ma2024agentboard, liu2023agentbench} designed to evaluate LLMs agents, their evaluation are often centered around investigating the models' performance under each specific application scenario and primarily measure task completion, which is hard to delve into the underlying capabilities that drive such outcomes. 
While existing benchmarks~\cite{mmlu:hendrycks2020measuring, yue2023mmmu, ma2024agentboard, liu2023agentbench} evaluate LLM agents by focusing on specific application scenarios and task completion, they struggle to reveal the underlying capabilities driving these outcomes.
For example, as shown in Fig. \ref{fig:error-type-example}, when an LLM encounters a complex math problem, multiple capabilities are required to solve it. By emphasizing task completion, existing benchmarks often obscure whether a failure stems from a lack of comprehension, reasoning, or calculation error. Consequently, these evaluation methods blur the distinctions between different types of failures, hindering our understanding of where the error originates from and limiting our ability to gain deeper insights into the model's capabilities and make targeted improvements.
% Additionally, many tasks in existing evaluation benchmarks require considerable effort to set up the environments, such as Digital Card Game, House-Holding, and Web-Shopping in AgentBench~\cite{liu2023agentbench}. Each of these tasks needs to be set up individually, making a thorough evaluation both expensive and challenging. 
Additionally, some tasks in existing benchmarks require considerable effort to set up the environments, making a thorough evaluation both expensive and challenging. 
Furthermore, we observe that tasks, especially interactive ones, are sometimes less stable and reproducible due to the stochasticity of the environment feedback during the evaluation process. This randomness can make it difficult to obtain consistent evaluation results and draw solid conclusions.

To address such limitations, we introduce the Massive Multitask Agent Understanding (MMAU) benchmark. We develop MMAU by identifing five essential capabilities: \textcolor{orange}{Understanding}, \textcolor{orange}{Reasoning}, \textcolor{orange}{Planning}, \textcolor{orange}{Problem-solving} and \textcolor{orange}{Self-correction} across five domains: \textcolor{teal}{Tool-use}, \textcolor{teal}{Directed Acyclic Graph (DAG) QA}, \textcolor{teal}{Data Science $\&$ Machine Learning coding}, \textcolor{teal}{Contest-level programming}, and \textcolor{teal}{Mathematics}. 
As a result, MMAU comprises a total of 3,220 distinct prompts gathered from diverse data sources. These include our in-house human annotations for tool-use, as well as rewritten and curated prompts from open-source datasets such as CodeContest~\cite{codecontest:li2022competition}, Kaggle~\cite{kaggle:jim_plotts_megan_risdal_2023}, and DeepMind-Math~\cite{deepmind-math:saxton2019analysing}. Based on this dataset, we designed 20 tasks across 64 subjects, offering a comprehensive benchmark.
To avoid the complexities of environment setup and issues of unreliability, all tasks in MMAU are performed on our 3K static dataset to eliminate potential issues related to environment instability.
% We also acknowledge that interactive evaluation is necessary. MMAU does not aim to replace them but rather to complement them by addressing the issues mentioned above. 
% % By offering a reliable and straightforward offline alternative, we hope to bridge the research gap. 
% Developing and providing a more stable and easy-to-use benchmark for interactive evaluations is valuable and warrants further studies.

We comprehensively evaluate 18 models on MMAU, which include both API-based commercial models and open-source models. In addition to conventional overall comparisons and evaluations tailored to specific application scenarios, we also study the varying capabilities across different models (Figure~\ref{fig:mmau_combined_eval_charts}). 
From our study utilizing MMAU, thorough analysis and insightful analysis arise (Sec.~\ref{sec:discussion}). 
% We identify the relationship between capabilities and domains (Sec.~\ref{sec:discussion}). For example, while it is commonly recognized that \textcolor{orange}{reasoning} is critical for math and coding, we also find that for complex and long-trace conversational tasks, \textcolor{orange}{planning} and strong \textcolor{orange}{understanding} capabilities are essential to accurately capture user intent and complex dependencies.
% Additionally, we detect model limitations that are not reflected in existing academic benchmarks. For example, we find that Phi-3~\cite{abdin2024phi-3} does not perform well in two coding-related domains on MMAU, which contrasts with its reported performance (Sec~\ref{sec:discussion}). When examining tasks specifically designed for decomposed capabilities, we observe the significant role of \textcolor{orange}{planning} for various models across different domains, which is not straightforward for detecting through standard evaluation protocols.
We envision MMAU as a valuable benchmark that not only yields significant observations but also equips the community with deeper insights. Our contributions are summarized as:
\begin{itemize}[leftmargin=*]
    \item We offers evaluations from both application scenarios and fundamental capabilities, providing a comprehensive framework for understanding the strengths and limitations of LLM agents.
    \item The evaluation process on MMAU is straightforward and unified on a static dataset, avoiding the instability issues that may arise from interactive evaluations and thus ensuring reliable results.
    \item We release our evaluation dataset and scripts, aiming to set a new standard for performance assessment in the AI landscape.
\end{itemize}

We also acknowledge that interactive evaluation is necessary. MMAU does not aim to replace them but rather to complement them by addressing the issues mentioned above. 
Developing and providing a more stable and easy-to-use benchmark for interactive evaluations is valuable and warrants further studies.

\section{Related Work}\label{sec:related_work}

\noindent \textbf{LLM-based Genelist Agents}
Researchers have proposed various generalist agent frameworks to create AI assistants that can understand and execute any instruction from users.
One pioneering work is Auto-GPT \citep{gravitasauto}, which uses a language model as an AI agent that can break down goals into actionable steps with the aid of auxiliary tools. Integrating language models into multi-agent collaboration systems \citep{chen2023agentverse} is a cutting-edge research area. Frameworks like AutoGen \citep{wu2023autogen}, LangChain \citep{langchain2022}, Camel \citep{li2023camel}, AGENTS \citep{zhou2023agents}, AutoAgents \citep{chen2024autoagents}, and XAgent \citep{xagent2023} have explored different approaches to enable communicative agents to collaborate autonomously, facilitate practical applications, ensure control and customization, dynamically generate specialized agents, and manage complex tasks effectively.

% \noindent \textbf{Agents for Coding}
% The application of these LLM-based autonomous agents has gained significant traction in software development due to their potential to streamline and optimize coding processes. SWE-Agent \citep{yang2024sweagent} integrates language models into the software development lifecycle, providing automated solutions for coding, testing, and maintenance tasks, thereby improving development efficiency and effectiveness. MetaGPT \citep{hong2023metagpt} employs language models to facilitate dynamic role adaptation and real-time problem-solving in collaborative multi-agent systems. ChatDev \citep{qian2023communicative} automates the entire software development lifecycle by mimicking the traditional waterfall model, encompassing design, coding, testing, and documentation phases.
% AutoCodeRover \citep{zhang2024autocoderover} combines language models with sophisticated code search capabilities to autonomously solve GitHub issues related to bug fixes and feature additions. AgentCoder \citep{huang2024agentcoder} employs a multi-agent framework to iteratively generate, test, and refine code, enhancing the effectiveness of code generation through thorough testing and feedback. CodeAct \citep{wang2024executable} uses executable Python code to interact with code interpreter and improve the action performance of LLM agents, outperforming traditional formats like JSON and text in various benchmarks.

\noindent \textbf{Agent Benchmarks}
The need for rigorous benchmarks also arises in response to the surge in LLM-based agents. Varying benchmarks are created to gauge agent capability on tasks inspired by real-world use cases. The Berkeley Function Calling Leaderboard \citep{berkeley-function-calling-leaderboard}, NexusRaven V2 Function Calling Benchmark \citep{nexusraven}, ToolBench \citep{qin2023toolllm}, StableToolBench \citep{guo2024stabletoolbench}, and API-BLEND \citep{Basu2024APIBLENDAC} seek to evaluate the capability of LLM agent to plan and perform function calls. Webshop \citep{yao2022webshop}, WebArena \citep{zhou2023webarena}, Mind2Web \citep{deng2023mind2web}, MiniWoB++ \citep{humphreys2022data}, and VisualWebArena \citep{koh2024visualwebarena} focus on the agent's ability to browse and interact with a web environment. A line of benchmarks consider the universal presence of user interface (UI) and envision UI automation agents, including PixelHelp \citep{seq2act}, MetaGUI \citep{sun2022meta}, MoTIF \citep{burns2021mobile}, AITW \citep{rawles2023android}, and OmniACT \citep{kapoor2024omniact}. SWE-bench \citep{jimenez2024swebench} tests agent capability to solve real-world software engineering problems. 

While each benchmark tends to focus on a specific application, a generalist agent should be able to perform well on a wide range of tasks. AgentBench \citep{liu2023agentbench} consolidates tasks covering coding, game, and math into a single systematic benchmark.
AgentBoard \citep{ma2024agentboard} evaluates agent capabilities under web browsing, tool use, embodied AI, and game domains. However, both benchmarks require containerized environments to run and need involved effort to implement new tasks. As a result, simply stacking tasks would lead to diminishing returns in comprehensiveness. MMAU takes a step back and considers a range of core agent capabilities. Based on the core capabilities, MMAU provides a range of tasks that are designed to produce decoupled metrics over the core capabilities. 

Table~\ref{tab:comparison_benchmarks} compares the supporting capabilities of various benchmarks. PlanBench~\citep{planbench:valmeekam2022large}, specifically designed for benchmarking planning capabilities in LLM agents, also supports disentangled evaluation of reasoning and planning. AgendaBoard~\citep{ma2024agentboard} provides a manually labeled subset for evaluating disentangled capabilities, focusing on interactive agents, such as spatial navigation and world modeling. In contrast, MMAU evaluates more fundamental and essential capabilities for LLM agents. Overall, MMAU offers a more comprehensive evaluation of fundamental capabilities in a disentangled manner.

% MMAU makes minimal assumptions about task design

% EWmbodied ALFWorld \citep{ALFWorld20}

% Highlight the challenging difficulty and multi-modal support of MMAU
\section{The MMAU Benchmark}
\label{sec:mmau}

\begin{table}[t]
\caption{Comparison of benchmarks in evaluating core capabilities of LLM agents. ``En.'' and ``Dis.'' represent entangled and disentangled, specifically. Understand.: understanding, Reason.: reasoning, Plan.: planning, Prob.-solv.: problem-solving, Self-corr.: self-correction, MM: multimodal grounding.}
\centering
\small
\resizebox{\textwidth}{!}{
\setlength{\tabcolsep}{3pt} % reduce space between columns
\newcolumntype{C}[1]{>{\centering\arraybackslash}p{#1}} % defining a new centered column type
\begin{tabular}{C{2.0cm}*{10}{C{1.0cm}}} % adjust the width as necessary
\toprule
Benchmarks  & \multicolumn{2}{c}{Understand.} & \multicolumn{2}{c}{Reason.} & \multicolumn{2}{c}{Plan.} & \multicolumn{2}{c}{Prob.-solv.} & \multicolumn{1}{c}{Self-corr.} & \multicolumn{1}{c}{MM} \\
            & En. & Dis. & En. & Dis. & En. & Dis. & En. & Dis. & & \\
\midrule
AgentBench~\cite{liu2023agentbench}  & \textcolor{green}{\ding{52}} & \textcolor{red}{\ding{56}} & \textcolor{green}{\ding{52}} & \textcolor{red}{\ding{56}}  & \textcolor{green}{\ding{52}} & \textcolor{red}{\ding{56}}  & \textcolor{green}{\ding{52}} & \textcolor{red}{\ding{56}} & \textcolor{green}{\ding{52}}  & \textcolor{green}{\ding{52}} \\
AgentBoard~\cite{ma2024agentboard}  & \textcolor{green}{\ding{52}} & \textcolor{red}{\ding{56}}  & \textcolor{green}{\ding{52}} & \textcolor{red}{\ding{56}}  & \textcolor{red}{\ding{56}}  & \textcolor{green}{\ding{52}}  & \textcolor{green}{\ding{52}} & \textcolor{red}{\ding{56}} & \textcolor{green}{\ding{52}}  & \textcolor{red}{\ding{56}} \\
PlanBench~\cite{planbench:valmeekam2022large}   & \textcolor{green}{\ding{52}} & \textcolor{red}{\ding{56}}  & \textcolor{green}{\ding{52}}  & \textcolor{green}{\ding{52}}  & \textcolor{green}{\ding{52}} & \textcolor{green}{\ding{52}} & \textcolor{red}{\ding{56}}  & \textcolor{red}{\ding{56}}  & \textcolor{red}{\ding{56}}  & \textcolor{red}{\ding{56}}  \\
MMLU~\cite{mmlu:hendrycks2020measuring}        & \textcolor{green}{\ding{52}} & \textcolor{red}{\ding{56}} & \textcolor{green}{\ding{52}}  & \textcolor{red}{\ding{56}}  & \textcolor{red}{\ding{56}}  & \textcolor{red}{\ding{56}}  & \textcolor{green}{\ding{52}}  & \textcolor{red}{\ding{56}}  & \textcolor{red}{\ding{56}}  & \textcolor{red}{\ding{56}}  \\
MMMU~\cite{yue2023mmmu}        & \textcolor{green}{\ding{52}} & \textcolor{red}{\ding{56}}  & \textcolor{green}{\ding{52}} & \textcolor{red}{\ding{56}} & \textcolor{red}{\ding{56}}  & \textcolor{red}{\ding{56}}  & \textcolor{red}{\ding{56}}  & \textcolor{red}{\ding{56}}  & \textcolor{red}{\ding{56}}  & \textcolor{green}{\ding{52}} \\
MMAU        & \textcolor{green}{\ding{52}} & \textcolor{green}{\ding{52}} & \textcolor{green}{\ding{52}}  & \textcolor{green}{\ding{52}}  & \textcolor{green}{\ding{52}} & \textcolor{green}{\ding{52}} & \textcolor{green}{\ding{52}}  & \textcolor{green}{\ding{52}}  & \textcolor{green}{\ding{52}}  & \textcolor{green}{\ding{52}} \\
\bottomrule
\end{tabular}
}
% \caption{Comparison of benchmarks in evaluating core capabilities of LLM agents.}
\label{tab:comparison_benchmarks}
% \vspace{-1.5em}
\end{table}

To introduce MMAU, we will start with an overview of all included capabilities~\ref{sec:mmau_capabilities}. We will then provide detailed explanations of how each task was designed and how the dataset across different domains was constructed \ref{sec:dataset_construction}.

\subsection{Capabilities in MMAU}
\label{sec:mmau_capabilities}
Below, we introduce the capability definitions and the key tasks used to evaluate them. A complete task-capability mapping can be found in the Appendix~\ref{tab:capability_task_mapping}.

\textcolor{orange}{\textbf{Understanding}} is a fundamental capability required of an intelligent agent. 
% To achieve any task, an agent must comprehend the user's intent from instructions, follow complex instructions effectively, and parse various kinds of information from the knowledge base. 
In MMAU, we evaluate an agent's understanding in different aspects, including:
% \textbf{Complex instruction following}: Evaluates the model's ability to follow complex requirements (Math-Comprehend+ and CodeContest-ProblemParsing).
% \textbf{User intent understanding}: Assesses the model's skill in capturing the implicit underlying intent based on its understanding of the context, which may not always be explicitly stated (tool-use).
% \textbf{Statistics parsing}: Measures the ability to parse and understand information from statistics, such as databases (DS\&ML QA).
% \textbf{Visual grounding}: Tests the agent's ability to grasp multimodal perception (VG-QA).
\textit{complex instruction following}, 
\textit{user intent understanding},
\textit{statistics parsing}, and
\textit{visual grounding}.

% \textcolor{orange}{\textbf{Reasoning and Planning}} is an essential yet challenging capability for intelligent AI agents. It reflects an agent's thought process and ability to infer logically from complex factors. Although reasoning and planning has been recognized in many works, it is often described as a general ability, compounded with other skills such as problem-solving, which limits deep investigation. Without categorizing and decomposing them effectively, measuring it accurately and improving it further becomes challenging. 
% In MMAU, we address this challenge with the task \texttt{planner-shift}, designed to decompose reasoning and planning capabilities from other factors. Unlike standard end-to-end evaluations, \texttt{planner-shift} divides the solution generation into two stages. In stage 1, a \texttt{planner model} generates a high-level plan, providing a strategy to solve the given problem without hinting at the final solution. In stage 2, a \texttt{solver model} is given the original problem along with the plan to solve it. This approach isolates the planning and reasoning processes from problem-solving. To test planning and reasoning capabilities, we vary only the \texttt{planner model} while using the same \texttt{solver model}, ensuring that performance differences reflect the planning and reasoning capabilities. The task design diagram is shown in Figure~\ref{fig:planer-solver-shift-diagram}.

\begin{figure}
    \centering
    \includegraphics[width=.75\textwidth]{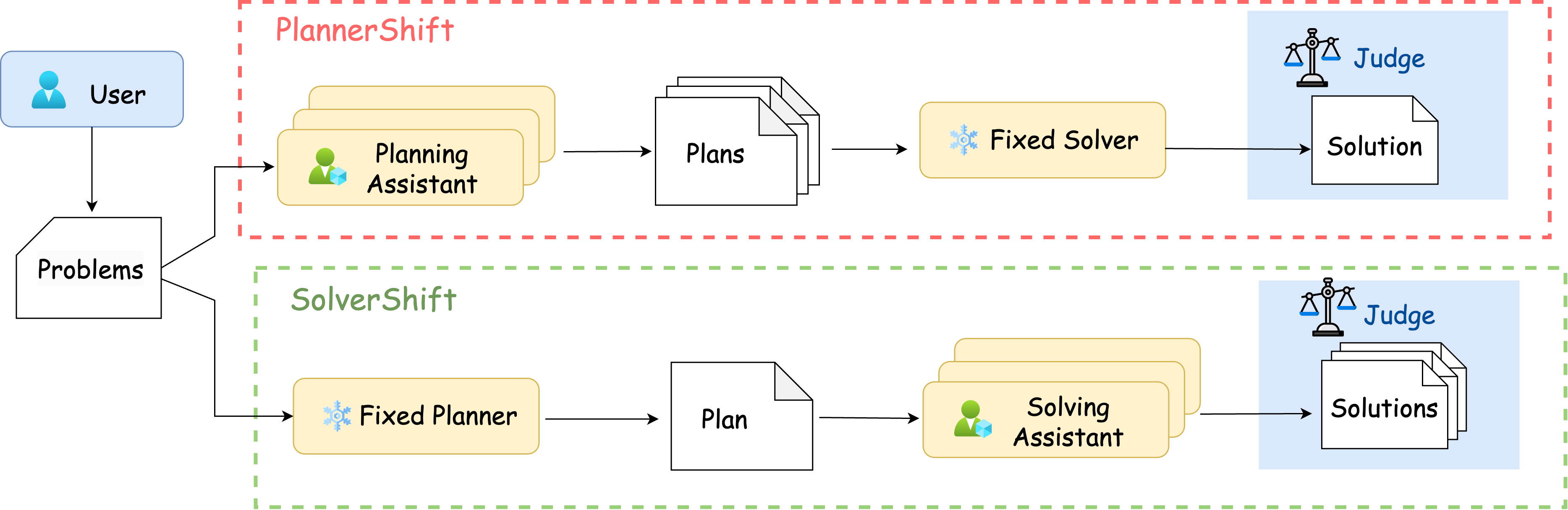}
    \caption{Construction of planner-shift task and solver-shift task.}
    \label{fig:planer-solver-shift-diagram}
    % \vspace{-2em}
\end{figure}

\textcolor{orange}{\textbf{Reasoning and Planning}} reflect an agent's thought process and ability to infer logically from complex factors. Although reasoning and planning has been recognized in many works, it is often described as a general ability, compounded with other skills, which limits deep investigation. 
In MMAU, we address this challenge with the task \texttt{planner-shift}, designed to decompose reasoning and planning capabilities from other factors. Unlike standard end-to-end evaluations, \texttt{planner-shift} divides the solution generation into two stages. In stage 1, a \texttt{planner model} generates a high-level plan, providing a strategy to solve the given problem without hinting at the final solution. In stage 2, a \texttt{solver model} is given the original problem along with the plan to solve it. This approach isolates the planning and reasoning processes from problem-solving. To test planning and reasoning capabilities, we vary only the \texttt{planner model} while using the same \texttt{solver model}, ensuring that performance differences reflect the planning and reasoning capabilities. The task design diagram is shown in Figure~\ref{fig:planer-solver-shift-diagram}.

% \textcolor{orange}{\textbf{Problem-solving}} focuses on measuring an agent's ability to successfully implement or execute a task, assuming it has already understood and well planned the strategy. This is a critical yet under-studied capability in agents, due to the complex factors in many domains. For example, in evaluating the success rate of a coding or tool-execution task, failures can arise from poor understanding, insufficient planning or reasoning, or errors in the final implementation, making it difficult to pinpoint the cause. 
% To address this, we designed a task called \texttt{solver-shift}, similar to \texttt{planner-shift}, which also performs a two-stage generation. However, unlike \texttt{planner-shift}, \texttt{solver-shift} keeps the \texttt{planner model} constant and varies only the \texttt{solver models} to reflect differences in problem-solving skills, as shown in Figure~\ref{fig:planer-solver-shift-diagram}.

\textcolor{orange}{\textbf{Problem-solving}} focuses on measuring an agent's ability to successfully implement or execute a task, assuming it has already understood and planned the strategy well.
To address this, we design a task called \texttt{solver-shift}, similar to \texttt{planner-shift}, which also performs a two-stage generation. However, \texttt{solver-shift} keeps the \texttt{planner model} constant and varies only the \texttt{solver models} to reflect differences in problem-solving skills, as shown in Figure~\ref{fig:planer-solver-shift-diagram}. In MMAU we use the tasks of \texttt{planner-shift} and \texttt{solver-shift} in domains of \textcolor{teal}{Contest-level coding}~\ref{sec:codecontest} and \textcolor{teal}{Math}~\ref{sec:math}.

\textcolor{orange}{\textbf{Self-correction}} is another core capability for an intelligent agent. It reflects the agent's ability to identify errors, learn from its environment and past behaviors, and correct itself to eventually overcome obstacles and achieve its task. In MMAU, we evaluate this capability by specifically designing self-correction tasks across different domains.

\subsection{Dataset Construction}
\label{sec:dataset_construction}
The construction of MMAU encompasses both breadth and depth of data, as illustrated in Table~\ref{tab:mmau_statistics}. 
% MMAU spans a wide array of domains and subjects, demonstrating its extensive breadth. Additionally, it includes challenging tasks such as contest-level coding problems, machine learning and statistics problems, as well as graduate-school level mathematics problems, and complex tool-use tasks, all of which exemplify its depth. 
Our dataset is constructed from heterogeneous sources: 1) our in-house tool-use data, used for tasks under \textcolor{teal}{tool-use} and \textcolor{teal}{DAG-QA}; 2) Kaggle~\cite{kaggle:jim_plotts_megan_risdal_2023} datasets, which we rewrote to design tasks for \textcolor{teal}{DS $\&$ ML } coding; 3) \textcolor{teal}{CodeContest}~\cite{codecontest:li2022competition}, used for tasks under contest-level coding; and 4) DeepMind-math~\cite{deepmind-math:saxton2019analysing}, used for \textcolor{teal}{math} tasks. MMAU involves curating and rewriting these data sources. In the following section, we will explain how we leveraged these source data to construct MMAU.

% \begin{wrapfigure}{r}{.6\textwidth}
%     \centering
%     \vspace{-1em}
%     \includegraphics[width=.59\textwidth]{figures/dag_qa_example.png}
%     \caption{\small {A DAG-QA example.}}
%     \label{fig:dag-qa-examples}
%     \vspace{-1em}
% \end{wrapfigure}

\subsubsection{Tool-Use} 
\label{sec:tool-use}
% In the context of this paper, we use the term "tool-use" and "function calling" interchangably. The standard function calling protocol involves the following steps.
% \begin{itemize}
%     \item The user sends a query to the agent model along with a list of potential functions including description of their purposes and parameters.
%     \item The agent responds with either natural language or appropriate use of functions.
%     \item In case of function call, the functions are invoked according to the agent either by the user or by the agent directly, and the result is submitted back to the agent model.
%     \item The agent can then conclude with the given information or continue the conversation with follow up questions or additional function calls.
% \end{itemize}
% We create an in-house dataset for tool-use with conversation trajectories of the above protocol. In particular, we select from a subset of popular RapidAPI functions and ask human annotators to create realistic scenarios with user query, ground truth function calls and actual function returns from making requests to the RapidAPI endpoints. 
% The dataset consists of 407 single-step and 256 multi-step tool-use conversations. Out of the 407 single-step tool use conversations, 224 require making parallel tool calls. We adapt these examples to for the following tasks.

%In the context of this paper, we use the term "tool-use" and "function-calling" interchangably. 
We curated an in-house dataset for tool-use with conversation trajectories following the standard tool-use (a.k.a. function-calling)  protocol.
% 1)  The user sends a query to the agent model along with a list of potential functions including a description of their purposes and parameters. 2) The agent responds with either natural language or appropriate function use. 3) In case of function-call, the functions are invoked according to the agent's instructions, either by the user or directly by the agent, and the result is submitted back to the agent model. 4) The agent can then conclude with the given information or continue the conversation with follow-up questions or additional function calls.
 We select from a subset of RapidAPI Hub~\footnote{https://rapidapi.com/hub} functions and ask human annotators to create realistic scenarios with user queries, ground truth function calls and actual function returns from the RapidAPI endpoints. 
In total, our in-house tool-use dataset consists of 409 single-step (\textbf{task:single-tool-use}) and 258 multi-step tool-use conversations (\textbf{task: multi-turn multi-tool-use}). Out of the 409 single-step tool use conversations, 225 require making parallel tool calls (\textbf{task: parallel-tool-use}). Figs.~\ref{fig:single-tool-use-samples}, \ref{fig:parallel-tool-use-samples} and \ref{fig:multi-tool-use-samples} show some of these examples. We adapt this dataset for the following tasks.

% \begin{wrapfigure}{r}{.6\textwidth}
%     \centering
%     \vspace{-1em}
%     \includegraphics[width=.59\textwidth]{figures/tool-use-examples.png}
%     \caption{\small {Examples from the tool use dataset.}}
%     \label{fig:tool-use-examples}
%     \vspace{-1em}
% \end{wrapfigure}

% \begin{figure}[h]
% \begin{subfigure}{0.5\textwidth}
% \includegraphics[width=.75\textwidth]{figures/tool-use-examples.png}
% \caption{\small{Examples from the tool use dataset.}}
% \end{subfigure}
% \begin{subfigure}{0.5\textwidth}
% \includegraphics[width=.75\textwidth]{figures/dag_qa_example.png}
% \caption{\small {A DAG-QA example.}}
% \end{subfigure}
% \end{figure}

\textbf{Task: Tool-use} Benchmarking agent tool-use following the standard protocol requires an interactive environment. To simplify the evaluation process, we instead evaluate the model's response at each assistant turn (i.e., where a function call is expected), conditioning on the ground-truth versions of all previous user or assistant turns. 
For evaluation, we check if the model's tool call matches that of the ground truth, i.e. calling the same function and the same parameters.
% To compare the predicted and the ground truth parameter values, we perform string normalization including stripping punctuation, white spaces and converting to lower case. In some cases where the parameter value can have open-ended, semantically equivalent forms, we define example-specific match rules to accommodate valid alternatives.

\textbf{Task: DAG QA} In this task, a user presents a set of requirements to which the LLM must respond by selecting and ordering a sequence of tool invocations from multiple choices provided. This design examines whether the model can identify the relevant tools and deduce the correct dependencies between them. The prompt enumerates the possible tool use orderings from which the LLM agent is asked to pick one, and the label is derived from the ground truth function call sequence.
% The reasoning and planning benchmarks differ only in the prompt, where for reasoning, the agent is requested to "elaborate on the thought process and reasoning", while for planning, the agent is requested to "be concise with a response in the format Chosen Plan: N". 
For example, we transform the multi-step example in Figure \ref{fig:parallel-tool-use-samples} into a DAG QA task as shown in Figure \ref{fig:dag-qa-examples}

\textbf{Task: Tool-use Self-correction} 
% From the single-step function calling execution dataset described above, we derive two classes of examples to test the model's self-correction ability:
% \begin{itemize}
%     \item \emph{temporary error} simulates a tool that is temporarily unavailable. From the ground truth messages [user query, tool call, tool response], we substitute a random temporary error (e.g. 429: too many requests, 504: gateway timeout) in lieu of the tool response, and we expect the model to retry the tool call on the next turn.
%     \item \emph{incorrect call} simulates a previous tool call or response containing an error. We mutate the ground truth tool call to be incorrect by changing the arguments or the function name, issue the modified call, and save the updated tool response. Given the message history of [user query, mutated call, updated tool response], we expect the model to issue the correct tool call next turn.
% \end{itemize}
% The evaluation metric is the exact match accuracy of function name and arguments against the ground truth. To encourage models to retry, we prepend the following to the first turn user message:
% \begin{verse}
% system \\
% A conversation between a user and a helpful assistant. The assistant can choose to directly generate text or make function calls to help with user queries. Retry the call if it does not succeed or if there are errors in the previous calls or responses. \\
% user
% \end{verse}
From the tool-use dataset described above, we derive two classes of errors to test the model's self-correction ability: \\
\texttt{Temporary error} simulates a tool that is temporarily unavailable. From the ground truth messages [user queries, tool-calls and tool responses], we substitute a random \texttt{temporary error} (e.g. \textit{``429: too many requests'', ``504: gateway timeout''}) in place of the tool response. \\
\texttt{Incorrect call} simulates a previous tool call or response containing an error. We mutate the ground truth tool-call to be incorrect by changing the arguments or the function name, issue the modified call, and save the updated tool response. Given the message history of [user queries, mutated calls, updated tool responses], the model is expected to retry with the correct call. \\
The evaluation metric is the exact match accuracy of the function name and arguments against the ground truth. 
% To encourage models to retry, we prepend the following system message to the user's first turn message:
% \begin{figure}[h!]
%     \centering
%     \vspace{-1em}
%     \includegraphics[width=.95\textwidth]{figures/retry_message.png}
%     % \caption{Caption}
%     \label{fig:enter-label}
%     \vspace{-1.5em }
% \end{figure}

% \begin{verse}
% system \\
% A conversation between a user and a helpful assistant. The assistant can choose to directly generate text or make function calls to help with user queries. Retry the call if it does not succeed or if there are errors in the previous calls or responses. \\
% user
% \end{verse}

% ================== coding ======================

\begin{wrapfigure}{r}{.4\textwidth}
    \centering
    % \vspace{-2em}
    \includegraphics[width=.42\textwidth]{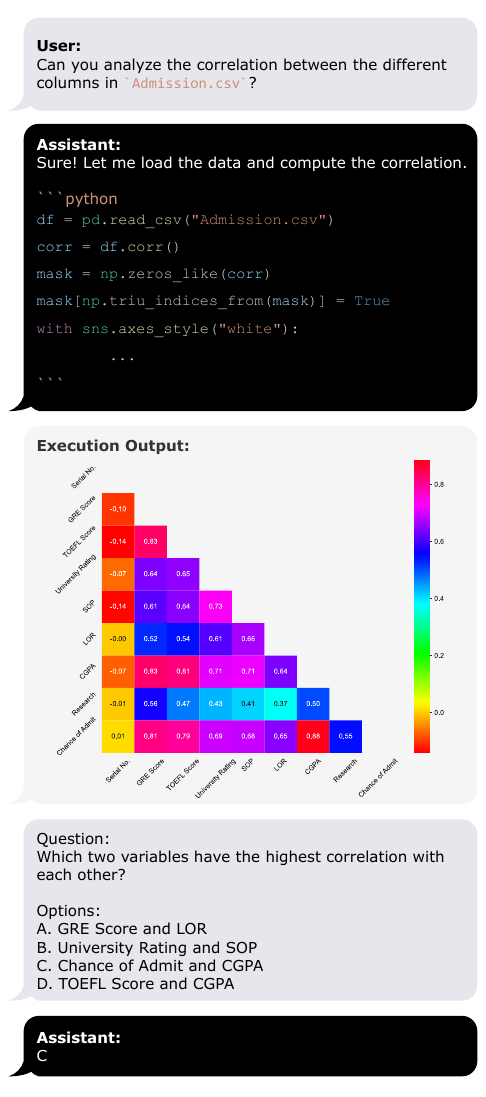}
    \caption{\small {A Multi-turn coding and QA  example for Data science and Machine learning.}}
    \label{fig:kaggle-example}
    \vspace{-2em}
\end{wrapfigure}

\subsubsection{Data Science and Machine Learning}

We leverage the Meta Kaggle Code dataset~\cite{kaggle:jim_plotts_megan_risdal_2023} and curate 28 Python notebook-style conversations, with 123 conversation turns. Each turn begins with a user request for code generation. Among all requests, 83 requests expect text-based outputs from code and 40 requests expect image outputs. Due to the open-ended nature of code generation, we created multiple-choice questions that require information from successful code execution to fully address, resulting in 207 text-based questions and 121 image-based questions. Figure \ref{fig:kaggle-example} shows an example turn with multiple choice questions. We report QA accuracy as the main metric and vary the combination of code model and QA model to produce different evaluation settings. 

% To represent all results on the same scale and reduce confusion, when QA model is not multimodal, we report its performance as $\mathrm{ratio\_of\_text\_based\_questions} * \mathrm{accuracy\_on\_text\_based\_questions} = (207 / 328) * \mathrm{accuracy\_on\_text\_based\_questions}$. We set the temperature to 0 for both code generation and QA.

% MMAU encompasses areas of data science and fundamental machine learning concepts. Our source data is derived from Kaggle datasets, which include notebook scripts from Kaggle dataset~\cite{kaggle:jim_plotts_megan_risdal_2023}. These topics range from statistics to basic machine learning and data science principles. We are leveraging GPT-4~\cite{gpt4:achiam2023gpt} to transform these notebooks into multi-turn dialogues to simulate interactions between an AI agent and a user. As a result, we have developed 29 multi-turn conversational notebooks, comprising a total of 125 turns. Of these, 87 turns are text-based, and 37 are image-based.

% Similar to the ``problem parsing'' task in coding contests, we have clearly separated coding from understanding. This allows us to evaluate these capabilities independently, without them being intertwined with other skills. Our tasks designed to assess understanding include both visual content and textual/statistical content. For each task, we create question-and-answer (QA) challenges that can be evaluated by measuring answer accuracy.

\textbf{Task: E2E Code Generation and QA}
% For the task focusing on textual and statistical content, we aim to evaluate the model's ability to understand user intent, parse statistics, and grasp basic machine learning concepts. We have curated 177 turns for which we developed corresponding question-and-answer (QA) tasks using GPT-4~\cite{gpt4:achiam2023gpt}. An example of this is depicted in Figure \ref{fig:kaggle-example}. Specifically, we select turns and employ GPT-4 to generate questions with multiple-choice options. These questions are designed to test solely the agent's ability to understand the information presented in the turn, without requiring additional skills such as programming.
% In this setting, we aim to gauge the overall capability of the model to understand user intent, generate code, and interpret output. the evaluated model is responsible for both code generation and QA.
In this setting, we aim to gauge the overall capability. the evaluated model is responsible for both code generation and QA.

\textbf{Task: Code Generation and GPT-4 QA }
% Additionally, MMAU is equipped to evaluate model proficiency in visual understanding, particularly with statistical visual content such as tables, graphs, and charts. To assess this capability, we have designed 107 question-and-answer (QA) tasks, where each question is paired with an image. For accurate responses, the model must correctly understand and extract specific numbers and features from the provided figures. We use answer accuracy as the evaluation metric for this task.
In this setting, we isolate the code generation capability of the model. After generating code from the evaluated model, we adopt a strong multimodal model (GPT-4~\cite{gpt4:achiam2023gpt}) to serve as control and perform QA based on code execution outputs.

\textbf{Task: QA from Oracle Code}
In this setting, we specifically focus on the textual and visual understanding proficiency of the model decoupled from code generation. We obtain oracle output by executing ground truth code implementation and then pass to the evaluated model to perform QA.

\textbf{Task: DS $\&$ ML Self-Correction}
This setting is similar to the E2E setting, however, whenever code execution fails, we use the execution error message to prompt an additional code generation turn. 

% \vspace{-0.5em}
\subsubsection{Contest-Level Coding} \label{sec:codecontest}
% \vspace{-0.5em}
% \subsubsection*{Contest-level coding}

% \begin{wrapfigure}{r}{.4\textwidth}
%     \centering
%     \vspace{-2.5em}
%     \includegraphics[width=.39\textwidth]{figures/codecontests_understand_example.png}
%     \caption{\small {An example of CodeContests ``ProblemParsing'' task to measure the agent's understanding capability.}}
%     \label{fig:codecontests-understand-problem}
%     \vspace{-1.5em}
% \end{wrapfigure}

For contest-level coding problems, we select 261 problems from the Valid and Test splits of the CodeContests dataset~\cite{codecontest:li2022competition} which includes competitive programming problems. We adapt these 261 problems for the following tasks.
% We select 261 valid problems from the Valid and Test splits of the CodeContests dataset~\cite{codecontest:li2022competition,alphacodium:ridnik2024code}, each contains a number of test cases, including public tests, private tests, and generated tests. Based on these 261 problems, we systematically evaluate various LLM models on the following types of tasks. 
% We focus on Python code generation. We adapt this dataset for the following tasks.
% Each problem is associated with a number of test cases, including public tests, private tests, and generated tests. 
% to evaluate the agent’s capabilities to solve competitive programming tasks. We further select the problems where the provided solutions can pass the test cases (to make sure that the test cases are valid). 
% We focus on Python code generation. This setup is following a recent work AlphaCodium~\cite{alphacodium:ridnik2024code}.

\textbf{Task: E2E Standard}
In this task, models are challenged with a variety of coding problems. The effectiveness of the solutions is measured by executing the code against all predefined test cases~\cite{codecontest:li2022competition}. All CodeContests results reported in this paper are based on pass@K (K=5) accuracy. 

\textbf{Task: Planner-shift and Solver-shift}
As introduced in Section~\ref{sec:mmau_capabilities}, we use thses two tasks to extensively measure the agent's capability in planning and problem-solving, respectively. We evaluate both of these tasks by generating K Python code solutions and verifying their pass rate.
% For \texttt{planner-shift}, the models being evaluated will be asked to generate a step-by-step plan in natural language. To evaluate the generated plan, we pass the problem description and the plan to a fixed strong solver model to generate K Python code solutions and verify their pass rate. 
% For \texttt{solver-shift}, We use a fixed planner model to pre-generate a plan for each problem, then the LLM being evaluated will be asked to generate K python code solutions. 

% \textbf{Task: Solver-shift}
% As a complementary task to  
% \textbf{Task: Planner-shift and Solver-shift}
% The task in the coding domain extends beyond mere implementation to emphasize strategic planning in code development. In this task, models are first asked to outline a plan, detailing the approach they intend to take to solve the problem. This preliminary stage focuses on assessing the model's ability to plan logically and efficiently without delving into actual coding.

% Following the initial planning stage, the model is provided with its own plan and tasked with translating it into functional code. The completed code is then evaluated against the same set of predefined test cases to ensure consistency and accuracy.

% Similarly, to distinctly assess planning capabilities within coding skills, we consistently use the same model as the coder in Stage II across all tasks, while varying the models that perform the planning in Stage I. This approach ensures that any differences in performance are solely reflective of the planning capabilities, without the confounding effects of coding and execution skills.

\textbf{Task: Problem Parsing}
Unlike the other tasks, this task does not require the model to write or execute any code. Instead, given a problem statement and associated test cases, the model is only tasked with predicting the outputs for these test cases. An example is shown in Figure~\ref{fig:codecontests-understand-problem}.
Our rationale is that if a model truly grasps the problem, including its complex instructions and user intent, it should be able to accurately predict the outputs based on its understanding alone. We use match accuracy as the evaluate metric.
% By separating the understanding from the execution phases in code development, this task aims to isolate the model's comprehension capabilities. 
% For this task, we use match accuracy as the evaluate metric, which is calculated based on the comparison between the model's output predictions and the ground truth answer for the provided test cases.

% \begin{figure}
%     \centering
%     \includegraphics[width=.8\textwidth]{figures/plan-solve-diagram_v2.png}
%     \caption{Construction of planner-shift task and solver-shift task.}
%     \label{fig:planer-solver-shift-diagram}
%     \vspace{-2em}
% \end{figure}

\textbf{Task: CodeContest Self-correction}
For the E2E standard task above, we collect the error messages of each candidate solution if it does not pass some test cases, including 4 types of errors: \textit{empty solution, compilation error, runtime error, and wrong outputs}. Then, we follow the setup as what we used in \texttt{Tool-use self-correction} (Sec.~\ref{sec:tool-use}) to append the error content as a feedback user message to the message list, and ask the model to try again. We still measure the pass@K metric by generating K candidates independently.
% For the problems where all the 5 candidates fail to pass, we will append the error content as a feedback user message in the message list, and ask the LLM to try again. To encourage the agent to retry, we also prepend an additional system instruction to the first user message before describing the problem. The evaluation metric is still pass@5, similar to Regular tasks.

% \vspace{-1em}
\subsubsection{Mathematics} \label{sec:math}
% \vspace{-0.5em}

% \begin{wrapfigure}{r}{.4\textwidth}
%     \centering
%     \vspace{-1em}
%     \includegraphics[width=.4\textwidth]{figures/math-understand-case1.png}
%     \caption{An example from for ``Comprehend+'' task.}
%     \label{fig:rewrite-math-problem}
%     \vspace{-1.5em}
% \end{wrapfigure}

The source data in the domain of math, derived from DeepMind-Math~\cite{deepmind-math:saxton2019analysing}, consists of 1,000 carefully curated math problems spanning 56 subjects, including calculus, geometry, statistics, and etc.

\textbf{Task: E2E Standard} We adhere to standard protocol by incorporating a Chain-of-Thought (CoT)~\cite{cot:wei2022chain} into the prompt to generate the answers end-to-end, and using accuracy as the evaluation metric.

\textbf{Task: Planner-shift and Solver-shift} As introduced in Sec.~\ref{sec:mmau_capabilities}, we use the two tasks to assess the model's planning and problem-solving abilities in a two-stage manner, avoiding confounding influences from other capabilities. The prompts used for each task can be found in the Appendix~\ref{sec:appendix_task_prompts}.

\textbf{Task: Comprehend+}. 
To better isolate and assess the understanding capability without excessive interference from other skills, we have devised a new task named \texttt{Comprehend+}. Our hypothesis is that problems that are straightforward mathematically but complex in their descriptions rely more heavily on understanding capabilities. To test this, we first selected a subset containing only the mathematically simpler problems, and then use an LLM to create new math problems that feature more complex descriptions or harder problem statements but retain the same underlying mathematical constructs from each data sample. A rewritten math problem example is shown in Fig. \ref{fig:math_comprehend_example}. After curation and verification, we finalize 676 newly created problems for \texttt{Comprehend+}.
Please refer to Appendix~\ref{sec:appendix_math_details} for details of dataset creation. 

\section{Evaluation}\label{sec:evaluation}
We comprehensively evaluate 18 models on MMAU. All evaluation model details are listed in Table~\ref{tab:eval_models}.
For easier reading, the main paper presents only the aggregated evaluation results. For the evaluation results over all 20 tasks, please refer to Appendix~\ref{sec:appendix_eval_results}.
% Our study utilizing MMAU offers a thorough and insightful analysis of LLM agents. We envision MMAU as a valuable benchmark that not only yields significant observations but also equips the community with deeper insights. Through this, we aim to facilitate advancements in the understanding and development of language models.

% \vspace{-1em}
\subsection{Domain-centric Evaluation and Analysis}
% \vspace{-0.5em}
\label{sec:domain_evals}

\begin{table*}[!tbp] % [!h]  % tbp
\caption{Domain-centric evaluation results. All values are reported as percentages ($\%$). Models that do not support tool use are labeled as N/A, while models not supporting multi-tool tool-use tasks are marked by *\protect\footnotemark. The bolded and underlined numbers indicate the 1st and 2nd highest performances in each category.}
\small
\centering
\scalebox{1.0}{
\begin{tabular}{llllll}
\toprule
\rowcolor{tbgray}  Model & Tool-use & DAG  & DS$\&$ML & CodeContests & Math \\
\midrule
GPT-4o~\cite{gpt-4o}  &  \underline{69.33}     &   \underline{77.38}    & \textbf{66.52}       & \textbf{31.80}     & \textbf{53.40}  \\
GPT-4-Turbo~\cite{gpt4} &  \textbf{75.13}  & \textbf{79.38}             & \underline{63.90}    & \underline{25.67}  & 38.57 \\
GPT-3.5-Turbo      & 66.63  & 28.60  & 35.79  & 10.34 & 25.00  \\
Gemini-1.5-pro~\cite{reid2024gemini} & 52.43 & 71.18 & 54.02 & 10.34 & \underline{39.70}  \\
Gemini-1.0-pro~\cite{team2023gemini} & 26.32 & 47.45  & 27.87  & 7.66 & 39.40   \\ 
Claude3 Opus~\cite{claude3}      & 62.39 & 73.61  & 59.45  & 15.33 & 37.40\\
Claude3 Sonnet~\cite{claude3}    & 51.92 & 57.65  & 52.44  & 10.34 & 26.80\\
Claude3 Haiku~\cite{claude3}    & 44.14  & 36.36  & 33.84  & 8.81 & 36.60 \\ 
\midrule
Mixtral-8x22B-v0.1~\cite{mixtral}        & N/A   & \textbf{72.51} & \textbf{31.10}   & \textbf{9.20}  & \textbf{50.00} \\
Mixtral-8x7B-v0.1~\cite{mixtral}         & N/A   & 30.82          & \underline{13.48}   & 1.92 & \underline{21.70}  \\
Mistral-7B-v0.2~\cite{jiang2023mistral}  & N/A   & \underline{34.15}   & 2.80 & 0.38 & 9.55 \\
Phi-3-mini4K-instruct~\cite{phi3} & N/A & 23.95 & 0.84 & 2.30 & 21.00 \\
Openfunctions-v2~\cite{gorilla-openfunctions-v2} & 26.53*  & 25.28    & 3.84   & \underline{8.05} & 15.75 \\
Hermes-2-Pro-Mistral-7B~\cite{Hermes-2-Pro-Mistral-7B} & \textbf{39.48} & 27.94 & 3.29 & 0.77  & 12.78 \\
Command R~\cite{command-r}        & 28.29* &  22.28 & 0.00      & 4.21 & 8.21 \\
% Command R+ & 0.55  & 1.02  & 16.71  & - & -    \\ 
LLama2-70B~\cite{touvron2023llama2} & N/A & 19.73  & 0.00 & 0.00 & 8.43 \\
Llama2-13B~\cite{touvron2023llama2} & N/A  & 17.96  & 0.00  & 0.00 & 4.10 \\
Llama2-7B~\cite{touvron2023llama2} & N/A  & 20.84 & 0.00  & 0.00  & 3.92 \\
\bottomrule
\end{tabular}}%
% \caption{Domain-centric evaluation results. }
\label{tab:domain_centric_eval_results}
\vspace{-1em}
\end{table*}

% For domain-centric evaluation, we chose tasks that rely solely on the evaluation model to solve them end-to-end and adhere to the standard setup used in the field. This ensures that the domain-centric results accurately reflect the most standardized evaluation outcomes, making it easier to compare performance across different benchmarks. Specifically, we use the evalution results of \texttt{E2E standard} task as domain evaluation results for \textcolor{teal}{Contest-level coding}, \textcolor{teal}{DS$\&$ML}, \textcolor{teal}{Math} and \textcolor{teal}{DAG-QA}. For \textcolor{teal}{Tool-use}, we report the weighted average of single-, parallel-, and multi-turn multi-tool-use tasks. The results are shown in Table~\ref{tab:domain_centric_eval_results}. 

 As shown in Table~\ref{tab:domain_centric_eval_results}, there is a clear performance gap between API-based commercial models and open-source models across all evaluation domains. Among the commercial models, the GPT-4 family (including GPT-4o and GPT-4) consistently outperforms other models. In math and contest-level coding, GPT-4o demonstrates a significant advantage. Additionally, Claude3-Opus and Gemini-1.5-pro perform reasonably well. 
% Command R performs reasonably well on tool-related tasks but falls behind other commercial models in coding and math domains. This could be because Command R is primarily trained to target agent, tool-use, and RAG (retrieval-augmented generation) capabilities, which may result in less balanced training for general domain tasks. 
While for open-source models, a significant number do not support tool-use. Among those that do, Hermes-2-Pro-Mistral-7B demonstrates strong tool-use performance.
For models that do not support tool use, Mixtral-8x22B-Instruct-v0.1 performs surprisingly well in math and DAG-QA, demonstrating its strong reasoning and planning capability. Additionally, Phi-3 performs well in math considering its model size. The Llama2 family, however, struggles with challenging coding tasks. 

% \vspace{-0.5em}
\subsection{Capability-centric Evaluation and Analysis}
\label{sec:capability_evals}

As introduced in Sec.\ref{sec:mmau_capabilities}, we designed tasks to decompose core capabilities from standard evaluations, allowing MMAU to offer a unique dimension of evaluation. Each capability includes tasks spanning different domains. To provide overall capability-centric evaluation results for each model, we aggregate tasks under each capability using a weighted average. Detailed task-capability mappings and the calculation method can be found in Appendix~\ref{sec:appendix_exp_details}. The overall capability-centric evaluation results are shown in Table~\ref{tab:capability_cap_results}. 

Notably, for the capability of \textcolor{orange}{Understanding}, GPT-4o significantly outperforms other models, demonstrating its superior capability in handling long contexts, complex user instructions, and capturing (sometimes implicit) user intents. 
Additionally, GPT-4, Gemini-1.5-pro, and Claude3-Opus also exhibit reasonably strong understanding capabilities.
% When examining the MistralAI and Llama2 families, it is consistently observed that larger-scale models significantly boost performance compared to their smaller counterparts. This suggests the importance of model capacity in learning from complex and lengthy user instructions. However, Llama2-13B performs on par with Llama2-7B, indicating that the performance improvement is not strictly linear and that there may be some emergent effects at play.
For the capabilities of \textcolor{orange}{Reasoning} and \textcolor{orange}{Planning}, the GPT-4 family shows the strongest performance.
When examining the capability of \textcolor{orange}{Problem-solving}, the performance gap is not significantly large. This trend suggests that when provided with "oracle" plans, solving a task may be less challenging. While models' problem-solving capabilities vary, most can perform these tasks reasonably well, indicating that this capability may be more universally achievable among different models.
On the contrary, for \textcolor{orange}{Self-correction}, we observe a significant gap among models. Among open-source models, aside from Mixtral-8x22B, others do not seem to possess the skill to reflect on and correct their own errors effectively. These evaluation results highlight that self-correction is a critical capability needing further research and development to advance the field.

\footnotetext{The current reported and documented prompt templates of these models do not support multi-tool execution. However, it can still be possible that models can call multiple tools with proper adaptation.}

\begin{table*}[!tbp] % [!h]  % tbp
    \caption{Capability-centric evaluation results. All values are reported as percentages ($\%$). Models that do not support tool use are labeled as N/A, while models not supporting multi-tool tool-use tasks are marked by *\protect\footnotemark. The bolded and underlined numbers indicate the 1st and 2nd highest performances in each category.}
\small
\centering
\resizebox{1.0\textwidth}{!}{
% \scalebox{0.9}{
\begin{tabular}{lccccccc}
\toprule
 \rowcolor{tbgray}  & \multicolumn{2}{c}{Problem-Solving} &  &  &  & \multicolumn{2}{c}{Self-correct} \\
 \rowcolor{tbgray} \multirow{-2}{*}{Model} &   w/o Tool-Use        &  w/ Tool-Use        & \multirow{-2}{*}{Understanding} & \multirow{-2}{*}{Reasoning} &     \multirow{-2}{*}{Planning} &     w/o Tool-Use       &       w/ Tool-Use    \\
\midrule
GPT-4o~\cite{gpt-4o} & \textbf{56.12} & \textbf{61.11} & \textbf{60.63} & \underline{50.47} & \underline{47.90} & \textbf{43.65} & \underline{51.56} \\
GPT-4-Turbo~\cite{gpt4} & 48.07 & \underline{58.29} & \underline{49.78} & \textbf{50.88} & \textbf{49.59} & \underline{40.86} & \textbf{51.86} \\
GPT-3.5-Turbo & 42.84 & 51.83 & 30.78 & 29.38 & 28.27 & 21.23 & 32.38 \\
Gemini-1.5-pro~\cite{reid2024gemini} & 49.04 & 50.32 & 47.63 & 36.28 & 33.77 & 34.69 & 36.66 \\
Gemini-1.0-pro~\cite{team2023gemini} & \underline{52.84} & 42.82 & 37.05 & 39.88 & 37.31 & 15.96 & 10.91\\ 
Claude3 Opus~\cite{claude3}      & 49.98 & 54.67 & 49.03 & 44.10 & 38.84 & 38.47 & 44.16 \\
Claude3 Sonnet~\cite{claude3}    & 40.47 & 44.79 & 40.22 & 36.57 & 36.92 & 31.01 & 37.05 \\
Claude3 Haiku~\cite{claude3}     &  39.77 & 41.42 & 36.85 & 42.09 & 42.09 & 20.44 & 30.01 \\
\midrule
Mixtral-8x22B-v0.1~\cite{mixtral}    &  \textbf{49.04} & N/A & \textbf{44.39} & \textbf{44.92} & \textbf{38.02} & \textbf{18.00} & N/A \\
Mixtral-8x7B-v0.1~\cite{mixtral}      & \underline{33.50} & N/A & \underline{27.98} & \underline{27.57} & \underline{30.67} & \underline{8.26} & N/A \\
Mistral-7B-v0.2~\cite{jiang2023mistral}       & 21.87 & N/A & 9.01 & 21.45 & 21.22 & 1.93 & N/A \\
Phi-3-mini4K-instruct~\cite{phi3} & 22.94 & N/A & 14.92 & 28.33 & 27.45 & 2.04 & N/A \\
Openfunctions-v2~\cite{gorilla-openfunctions-v2}  & 20.89 & 29.43* & 11.20 & 23.30 & 24.76 & 2.61 & \textbf{22.77*} \\
Hermes-2-Pro-Mistral-7B~\cite{Hermes-2-Pro-Mistral-7B} & 29.26 & \textbf{33.12} & 16.54 & 25.95 & 24.96 & 2.04 & 2.99 \\
Command R~\cite{command-r}      & 22.17 & 31.30* & 19.47 & 22.72 & 22.95 & 0.35 & 18.84* \\
LLama2-70B~\cite{touvron2023llama2} & 24.23  & N/A & 6.32 & 15.30 & 14.55 & 0.34 & N/A \\
Llama2-13B~\cite{touvron2023llama2} & 12.86  & N/A & 2.69 & 13.39 & 13.15 & 0 & N/A \\
Llama2-7B~\cite{touvron2023llama2} & 15.92 & N/A & 2.38 & 14.25 & 13.26 & 0 & N/A \\
% Llama-2-7B~\cite{touvron2023llama2} &
% Llama-2-13B & 
% Llama-2-70B &
% Model & Understanding & Reasoning  & Planning & Problem-solving & Self-correct \\
% \midrule
% GPT-4o-2024-05-13  &       &       & 66.52       & 31.80     &   \\
% GPT-4-Turbo        & 1.26  & 3.48  & 63.90  & 25.67 & 54.04  \\
% GPT-3.5-Turbo      & 1.26  & 1.26  & 35.79  & 10.34 & 25.00  \\
% Gemini-1.0-pro~\cite{team2023gemini} & 27.87  & 7.66  & 0.91  & 0.80 & 39.40   \\ 
% GeminiPro-1.5-pro & 0.83 & 3.10 & 54.02 & 10.34 & 39.70  \\
% Claude3 Haiku     &  &  & 33.84  & 8.81 & \\ 
% Claude3 Sonnet    & &   & 52.44  & 15.41 & \\
% Claude3 Opus      & &   & 59.45  & 15.33 & \\
% Command R         & &   & 0.00      & 4.21 & \\
% Command R+ & 0.55  & 1.02  & 16.71  & - & -    \\ \midrule
% Openfunctions-v2  &  &     & 3.84   & 8.05 &15.75 \\
% Hermer-2-Pro-Mistral-7B &  & & 3.29 & 0.77 & 12.78 \\
% Phi-3-mini4K-instruct & N/A & & 0.84 & 2.30 & 21.00 \\
% Mistral-7B       & N/A  &     & 2.80 & 0.38 & 9.55 \\
% Mixtral-8x7B      & N/A  &    & 13.48  & 1.92 & 21.70  \\
% Mixtral-8x22B-v0.1    & N/A   & 31.10  &  & 3.07  & 50.00 \\
% Llama-2-7B~\cite{touvron2023llama2} & N/A  &  & 0.00  & 0.00  & 3.92 \\
% Llama-2-13B & N/A  &  & 0.00  & 0.00 & 4.10 \\
% Llama-2-70B & N/A &   & 0.00 & 0.00 & 8.43 \\

\bottomrule
\end{tabular}}%
% \vspace{-2.5em}
\label{tab:capability_cap_results}
\end{table*}

% domain-centric eval 

% \vspace{-0.5em}
\section{Analysis and discussion}
\label{sec:discussion}
\textbf{How does planning impact the performance?} One interesting finding emerges from the results of our designed tasks, \texttt{Planer-Shift} and \texttt{Solver-Shift} on \textcolor{teal}{Math}. 
As shown in Table~\ref{tab:math_planner_solver_shift_eval}, we find that high-quality planning can boost performance for all models on Math. For example, Command R's performance increases from 8.21$\%$ to 33.33$\%$, and Llama-2-70B's from 8.43$\%$ to 32.10$\%$. Even already strong models, such as Mixtral-8x22B, saw improvement from 50$\%$ to 60.02$\%$. 
% Such trends were also observed in the domains of CodeContest.
Interestingly, using the model itself as the planner also improves performance, e.g. bumping GPT-4o from 53.4$\%$ to 61.2$\%$. This shows that explicitly instructing the model to first develop a high-level strategy and then solve the problem based on that strategy can be a promising approach to further enhance performance.
% On the task of \texttt{Planer-Shift}, where we used the same solver model but varied the planner models, we found a weak planner can significantly hurt the final performance. For example, performance dropped from 32.80$\%$ to 9.32$\%$ when we changed the planner from GPT-3.5-turbo to Llama-2-13B. 
% The full results can be found in Appendix~\ref{sec:appendix}.

\textbf{Do different capabilities present varying levels of difficulty for models to achieve?}
Our evaluation results reveal that different capabilities indeed present varying levels of difficulty for models to achieve. As what we mentioned in Sec.~\ref{sec:capability_evals}, \textcolor{orange}{problem-solving} capabilities exhibit a smaller performance gap among models, suggesting that problem-solving is a more universally achievable capability across different models. However, \textcolor{orange}{self-correction} capabilities present a significant challenge, with a notable performance gap observed among models, and many open-source models lacking effective self-correction skills. These findings suggest that while some capabilities like \textcolor{orange}{problem-solving} are more readily attained by current models, others, such as \textcolor{orange}{self-correction} and \textcolor{orange}{planning} pose greater challenges and are vital areas for future advancements in the field.

\textbf{Do balanced capabilities indicate the path to a generalist agent?}
From Figure~\ref{fig:mmau_combined_eval_charts}, we observe that strong models, such as the GPT-4 family, exhibit balanced performance across all capabilities, demonstrating their robust and versatile nature. This balance indicates that improvements in one area likely enhance performance in others, highlighting a high correlation and interdependence among these capabilities. Conversely, models that perform poorly in one capability tend to struggle across the board, suggesting underlying weaknesses in their architecture or training strategy.

% llama2 families vs. mistral families, model scale
\textbf{Are larger models always better?} Another interesting finding arises when comparing the MistralAI families and Llama-2 families. For the MistralAI models, we consistently observe performance gains with increasing model size across domains. However, this trend does not apply to the Llama-2 families. In code-related domains (DS $\&$ ML, CodeContest), all size variants of Llama-2 perform poorly. Surprisingly, in the DAG-QA domain, the Llama-2-7B model performs better than its larger counterparts. This observation is consistent with findings from AgentBench~\cite{liu2023agentbench}, further validating that training strategies and model architectures also influence the scaling law.

\section{Conclusion}\label{sec:conclusion}

In this paper, we introduce the Massive Multitask Agent Understanding (MMAU) benchmark. By evaluating models based on both application scenarios and fundamental capabilities, MMAU provides a comprehensive and in-depth test bed for reliable and thorough studies. By designing 20 tasks to decompose capabilities beyond standard evaluation benchmarks, MMAU offers more granular insights into the strengths and limitations of these models.

\textbf{Limitations and future work.} 
The current scope of MMAU, while broad, does however not encompass all possible domains relevant to LLM agents, such as interactive environments which are also critical yet challenging. Future iterations of MMAU should aim to include interactive tasks to provide a more holistic evaluation. This expansion will require the development of reliable, stable, and user-friendly interactive environments.
Moreover, as we expand to include more domains, it will be essential to incorporate additional capabilities such as retrieving, memorizing, sequential decision-making, etc. Our current approach to capability decomposition, though insightful, still faces challenges in disentangling compound capabilities. Future research should focus on developing more effective methods for decomposing and evaluating these capabilities to further refine the benchmark.

\textbf{Ethics and Societal Impacts.}
Research on LLM agents must consider potential ethical concerns and negative societal impacts. The MMAU benchmark aims to provide a thorough and transparent evaluation framework, but it is crucial to ensure that these evaluations do not inadvertently reinforce biases or propagate harmful content. We are also careful in detecting and mitigating any personally identifiable information or offensive content within our datasets and prompts.

% \newpage

\section*{Acknowledgement}
We acknowledge and thank Mark Lee who helped with our code review and release.

\bibliographystyle{unsrt}{\small
\bibliography{egbib}
}
\newpage
\appendix

\newpage

\section{Experiment and dataset details}
\label{sec:appendix_exp_details}

The key statistics of the MMAU dataset are presented in Tab.~\ref{tab:mmau_statistics}. A list of all evaluated models (commercial and open source) is provided in Tab.~\ref{tab:eval_models}.

\begin{table}[h!]
\caption{The key statistics of MMAU.}
    \centering
    % \small 
    \resizebox{\textwidth}{!}{
    \begin{tabular}{l|lcccccc}
    \toprule
                               & Source data  & subjects  & task             & prompts & turns  & Answer type  \\ \hline
    \multirow{4}{*}{Tool-use}  & \multirow{4}{*}{In-house} & \multirow{4}{*}{Sports, Health, Location, etc}  & single-tool. & 409  & N/A  & function-call \\
                               &                           &  & multi-tool.  & 258  & N/A  & function-call \\
                               &                           &  & DAG QA   & 695  & N/A  & multi-choice \\
                               &                           &  & self-correct   & 282  & N/A  & function-call \\ \hline
    \multirow{4}{*}{CodeContest} & \multirow{4}{*}{CodeContest~\cite{codecontest:li2022competition}} & \multirow{4}{*}{alg., datastr., etc} & standard & \multirow{4}{*}{261} & \multirow{4}{*}{N/A} & execution \\
                                 &  &  & problem parsing  &  &  & multi-choice \\
                                 &  &  & PlannerShift  &  &  & execution \\
                                 &  &  & Solvershift   &  &  & execution \\ \hline 
    \multirow{3}{*}{ML$\&$DS}  & \multirow{3}{*}{Kaggle\cite{kaggle:jim_plotts_megan_risdal_2023}}  & \multirow{3}{*}{DS, ML, Visual} & textual QA  & 207  & N/A  & multi-choice \\ 
                               &   &     & code generation & 123 & N/A  & execution \\
                               &   &     & visual QA & 121  & N/A  & multi-choice \\ \hline 
    \multirow{4}{*}{Math}   & \multirow{3}{*}{DM-math\cite{deepmind-math:saxton2019analysing}}  & \multirow{4}{*}{calculus, geometry...} & standard      & \multirow{3}{*}{1K} & \multirow{4}{*}{N/A} & \multirow{4}{*}{math solution} \\ 
                            &                           &                      & PlannerShift & &  & \\
                            &                           &                      & SolverShift  &  &  & \\
                            &  synthetic                &                      & Comprehend+  & 676  &  &  \\ 
                                 
    \bottomrule
    \end{tabular}
    }
    % \caption{The key statistics of MMAU.}
    \label{tab:mmau_statistics}
\end{table}

\begin{table}[h!]
    \small 
    \centering
    \caption{Evaluation models in MMAU.}
    \begin{tabular}{l|l}
    \toprule
         \rowcolor{tbgray} API-based Commercial Models & Open-source Models \\ \hline
         GPT-4o-2024-05-13           & Mixtral-8x22B-Instruct-v0.1  \\
         GPT-4-turbo-2024-04-09      & Mixtral-8x7B-Instruct-v0.1  \\
         GPT-3.5-turbo-0125          & Mistral-7B-Instruct-v0.2  \\
         Gemini-1.5-pro-preview-0409 & Phi-3-mini-4k-instruct \\
         Gemini-1.0-pro              & gorilla-openfunctions-v2 \\
         Claude-3-opus-20240229      & Hermes-2-Pro-Mistral-7B \\
         Claude-3-sonnet-20240229    & c4ai-command-r-v01 \\
         Claude-3-haiku-20240307     & c4ai-command-r-rplus \\
                                     & Llama-2-70b-chat-hf \\ 
                                     & Llama-2-13b-chat-hf \\ 
                                     & Llama-2-7b-chat-hf \\
    \bottomrule
    \end{tabular}
    \label{tab:eval_models}
\end{table}

For our evaluations the open source models have been hosted via VLLM \footnote{\url{https://github.com/vllm-project/vllm}} on up to 8 NVidia A100 or H100 GPUs per model depending on the size. We have used the checkpoints from Huggingface for the open source models.

In order to have deterministic prediction we use a temperature of 0 and greedy search for all models and tasks.

\subsection{Tool-use}
Data construction protocol:
1)  The user sends a query to the agent model along with a list of potential functions including a description of their purposes and parameters. 2) The agent responds with either natural language or appropriate function use. 3) In case of function-call, the functions are invoked according to the agent's instructions, either by the user or directly by the agent, and the result is submitted back to the agent model. 4) The agent can then conclude with the given information or continue the conversation with follow-up questions or additional function calls.

Tool-use evaluation details:  To compare the predicted and the ground truth parameter values, we perform string normalization including stripping punctuation, white spaces and converting to lower case. In some cases where the parameter value can have open-ended, semantically equivalent forms, we define example-specific match rules based on regular expressions to accommodate valid alternatives.

\subsection{DAG QA} 
\begin{figure}
    \centering
    \includegraphics[width=.8\textwidth]{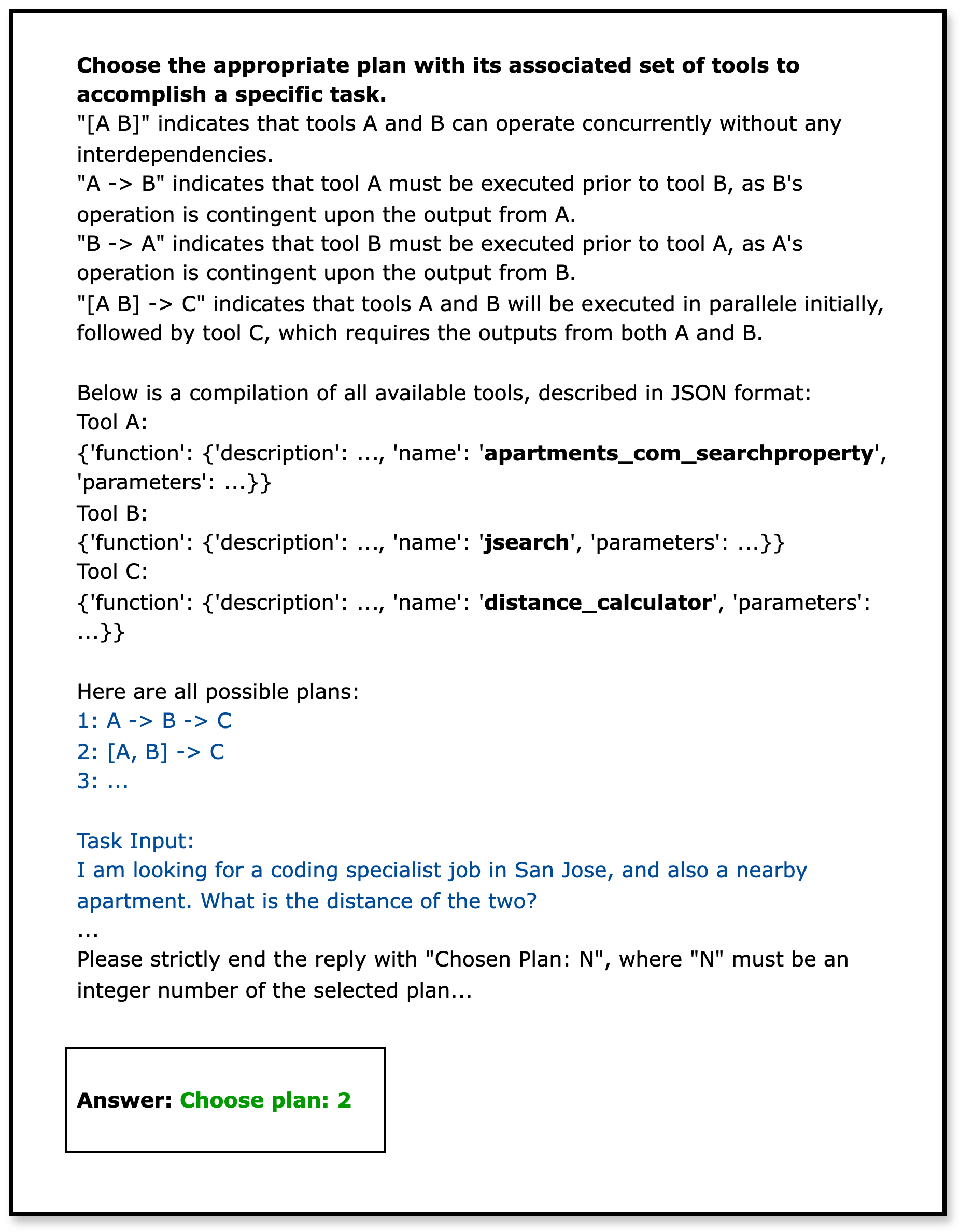}
    \caption{\small {A DAG-QA example.}}
    \label{fig:dag-qa-examples}
\end{figure}

Data construction protocol: The query includes a description of the task (to choose the appropriate plan), a list of potential functions (including a description of their purposes and parameters), an enumeration of all possible plans (all possible sequences in which to execute the tools), and the task input (what the task is that the plan is expected to solve). The task is formulated as a multiple choice task, where at the end of the query, the agent is asked to end the reply with which plan it chooses. At evaluation, the chosen plan is then extracted from the output by searching for a string match of the requested response format. An example prompt is illustrated in Figure~\ref{fig:dag-qa-examples}.

The reasoning and planning benchmarks differ only in the prompt, where for reasoning, the agent is requested to "elaborate on the thought process and reasoning", while for planning, the agent is requested to "be concise with a response in the format Chosen Plan: N".

\subsection{Self-correction tool-use}
To encourage models to retry, we prepend the following system message to the user's first turn message:
\begin{figure}[h!]
    \centering
    \vspace{-1em}
    \includegraphics[width=.95\textwidth]{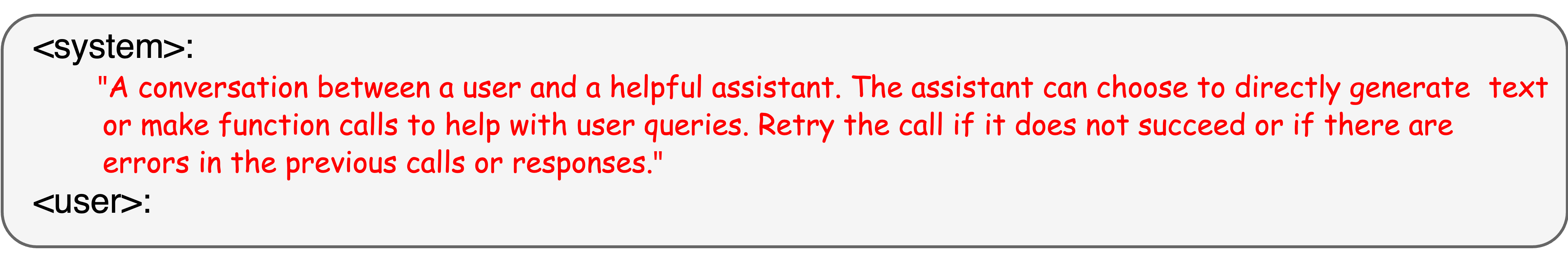}
    \caption{retry-message}
    \label{fig:retry_message}
    \vspace{-1.5em }
\end{figure}

\textbf{Kaggle}: 
To represent all results on the same scale and reduce confusion, when QA model is not multimodal, we report its performance as $\mathrm{ratio\_of\_text\_based\_questions} * \mathrm{accuracy\_on\_text\_based\_questions} = (207 / 328) * \mathrm{accuracy\_on\_text\_based\_questions}$. We set the temperature to 0 for both code generation and QA.

\subsection{Contest-level Coding}
\label{sec:appendix_codecontests_details}

% \textbf{Experiment Setup for CodeContests}

We select 261 valid problems from the Valid and Test splits of the CodeContests dataset~\cite{codecontest:li2022competition}, each contains a number of test cases, including public tests, private tests, and generated tests. 

\textbf{E2E Standard.}
The agent is asked to solve the problem by generating Python code solution.
The prompt contains a detailed description of the CodeContests problem with public test examples, as well as basic instructions on code formatting following the baseline prompt design in~\cite{alphacodium:ridnik2024code}. The generated code solution will be compiled and executed on all test cases, and is considered correct only if all test cases are passed. Following related works, we report the pass@5 score, which is the percentage of problems solved by generating 5 solutions per problem. We set temperature as 0.3 for all models.

\textbf{Problem Parsing.}
We design a ProblemParsing task for each CodeContests problem, which provides the agent with the problem description and public test examples, and asks the agent to directly infer the desired output for an unseen test case. To make the unseen test case relatively easy for parsing and computation, we select the shorted test case from the union of private test cases and generated test cases. We adopt a chain-of-thought prompt to help the LLM agent understand the problem (see Appendix~\ref{sec:appendix_task_prompts} for detailed prompts). We then compare the answer with the groundtruth output, and derive the accuracy over the entire dataset. For reproducibility, we set temperature as 0 for all models.

\textbf{Planner-shift.}
As introduced in \ref{sec:codecontest}, we divide the code generation process for CodeContests problems into two steps: generating a plan (i.e., planner) and generating code solutions (i.e., solver). We evaluate the planning capability of LLM agents by fixing the solver to be the same strong model. For each LLM model, we generate 1 plan with temperature 0, and use gpt-4-0125-preview as the solver to generate 5 code solutions (with temperature 0.3) and calculate the percentage of solved problems. 

\textbf{Solver-shift.}
SolverShift complements PlannerShift by freezing the planner and evaluating the problem-solving capability of various LLM models. Given 1 plan generated by gpt-4-0125-preview, we let LLM agents generate 5 code solutions and calculate the percentage of solved problems.

\textbf{Self-correction / Retry.}
We also evaluate the self-correct capability of LLM agents on the CodeContests problems. For the Regular task above, we collect the error messages of each candidate solution if it does not pass some test cases, including 4 types of errors: empty solution, compilation error, runtime error, and wrong outputs. For the problems where all the 5 candidates fail to pass, we will append the error content as a feedback user message in the message list, and ask the LLM to try again. To encourage the agent to retry, we also prepend an additional system instruction to the first user message before describing the problem. The evaluation metric is also pass@5.

\subsection{Math}
\label{sec:appendix_math_details}
\begin{figure}
    \centering
    \includegraphics[width=.8\textwidth]{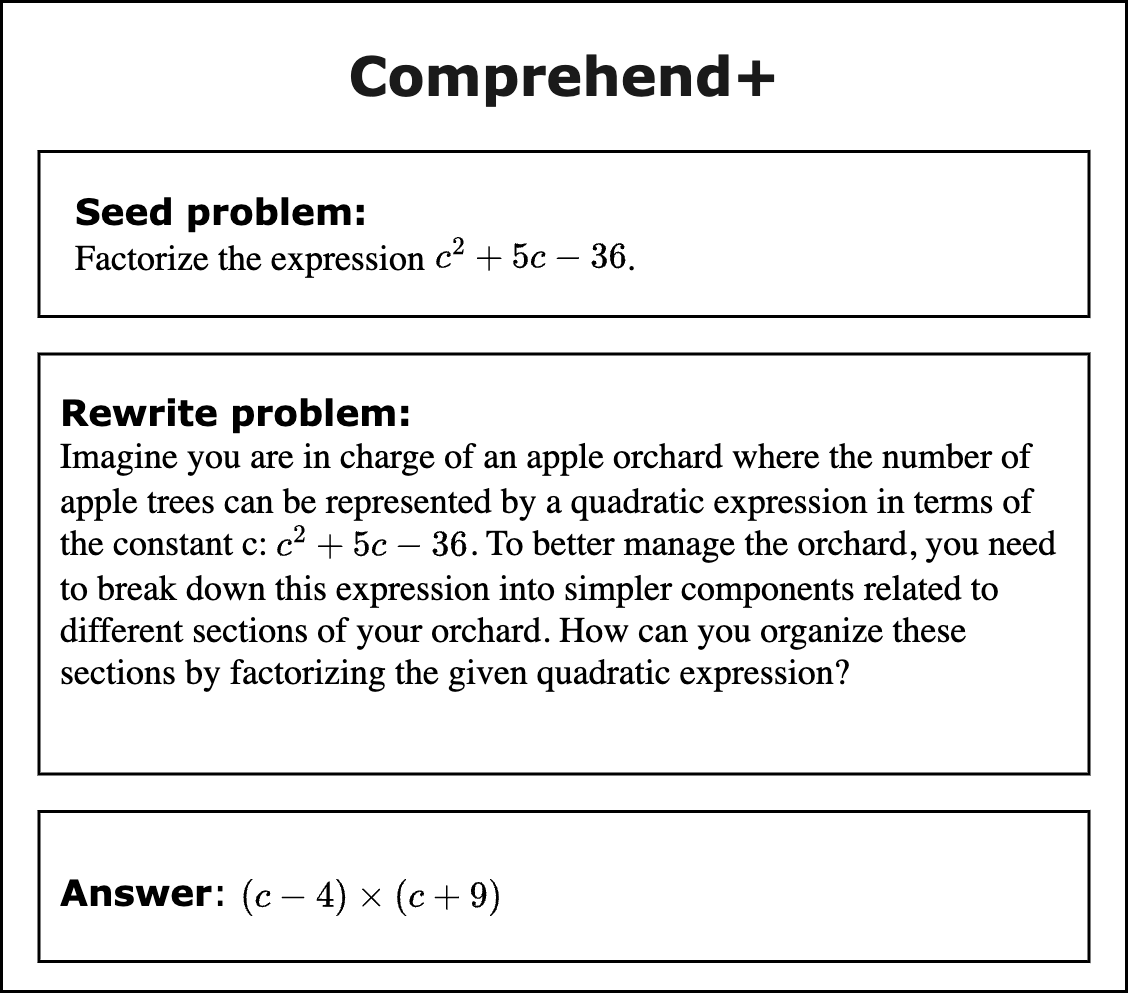}
    \caption{Example from MathComprehend+.}
    \label{fig:math_comprehend_example}
\end{figure}

Given a set of seed math problems $S$, we first employ a LLM as the judger $M_J$ to determine the difficulty level for each problem, resulting in a labeled math problem set $S':\{s_i^{d}\}$. We then select a subset containing only the mathematically simpler problems, i.e., $S'': \{s_i^d, d \leq \theta\}$. As a result we have a curated ``simple'' math problem set $S''$ (672 problems). Next, we use another LLM $M_w$ (GPT-4~\cite{gpt4:achiam2023gpt}) to create new math problems that feature more complex descriptions or harder problem statements but retain the same underlying mathematical constructs for each data sample $S''': \{ M_w(s_i^d), d \leq \theta \}$. A rewrite math problem example is shown in Fig. \ref{fig:math_comprehend_example}. Finally, to ensure that each rewritten problem in $S'''$ remains valid and has the same ground truth answer as $S$, we perform a data curation stage. As a result, we finalize the problem set $\mathcal{S^*}$ for the \texttt{Comprehend+} task.

\begin{table}[h]
    \centering
    \caption{Mapping of capabilities and tasks.}
    \label{tab:capability_task_mapping}
    \resizebox{1.0\textwidth}{!}{
    \begin{tabular}{>{\centering\arraybackslash}p{2cm}|>{\centering\arraybackslash}p{2.5cm}>{\centering\arraybackslash}p{2cm}>{\centering\arraybackslash}p{2.5cm}>{\centering\arraybackslash}p{2.5cm}>{\centering\arraybackslash}p{2cm}}
    \toprule
         \rowcolor{tbgray} Capability & Tool-use & DAG & DS\&ML & CodeContests & Math  \\
         \hline
         Problem-solving & Tool Execution & - & Code Generation and GPT-4 QA & Solver-shift & Solver-shift \\
         \hline
         Understanding & - & - & QA from Oracle Code & Problem Parsing & Comprehend+ \\
         \hline
         Reasoning & - & DAG QA w/ Reasoning & - & Planner-shift  & Planner-shift  \\
         \hline
         Planning & - & DAG QA w/o Reasoning & - & Planner-shift  & Planner-shift  \\
         \hline
         Self-correct & Self-correction & - & Self-correction & Self-correction & - \\
    \bottomrule
    \end{tabular}
    }
\end{table}

\clearpage
\lstset{
  basicstyle=\footnotesize\ttfamily,
  columns=fullflexible,
  breaklines=true,
  breakindent=0pt,
  xleftmargin=2em,
  xrightmargin=2em,
  frame=none,
  extendedchars=true,
  escapechar=@,
  literate={á}{{\'a}}1 {ã}{{\~a}}1 {é}{{\'e}}1 {£}{{\pounds}}1 {–}{{-}}1 {’}{{'}}1,
}
\lstset{frame=lines}

\section{Prompt templates}
\label{sec:appendix_task_prompts}

% \begin{itemize}
%     \item Tool-execution
%     \item DAG
%     \item self-correct
%     \item DS and ML
%     \item CodeContest
%     \item Math
% \end{itemize}

\subsection{Tool-use}

\begin{figure}[h!]
    \centering
    \includegraphics[width=\textwidth]{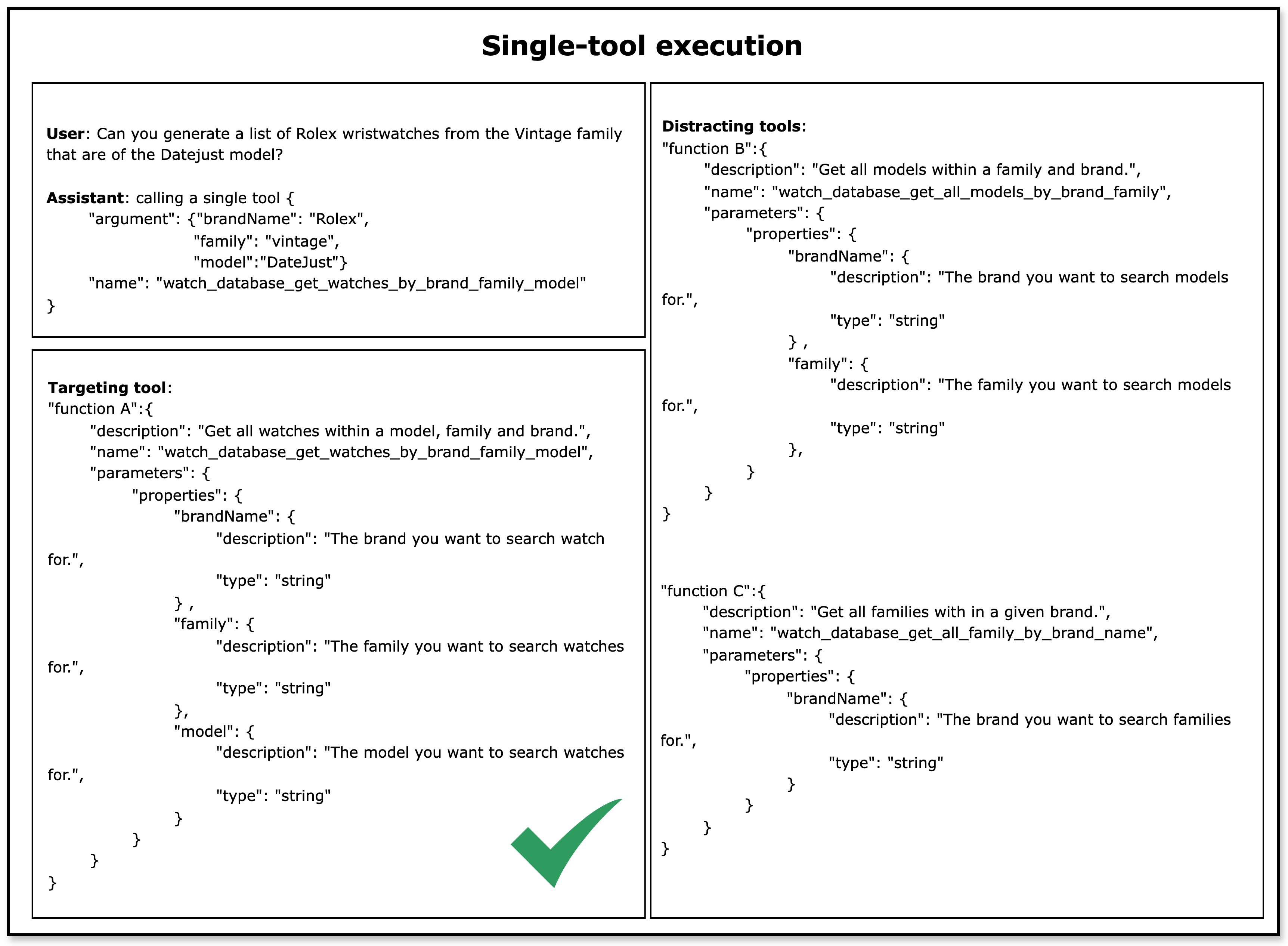}
    \caption{Example of single tool-use}
    \label{fig:single-tool-use-samples}
\end{figure}

\begin{figure}[h!]
    \centering
    \includegraphics[width=.85\textwidth]{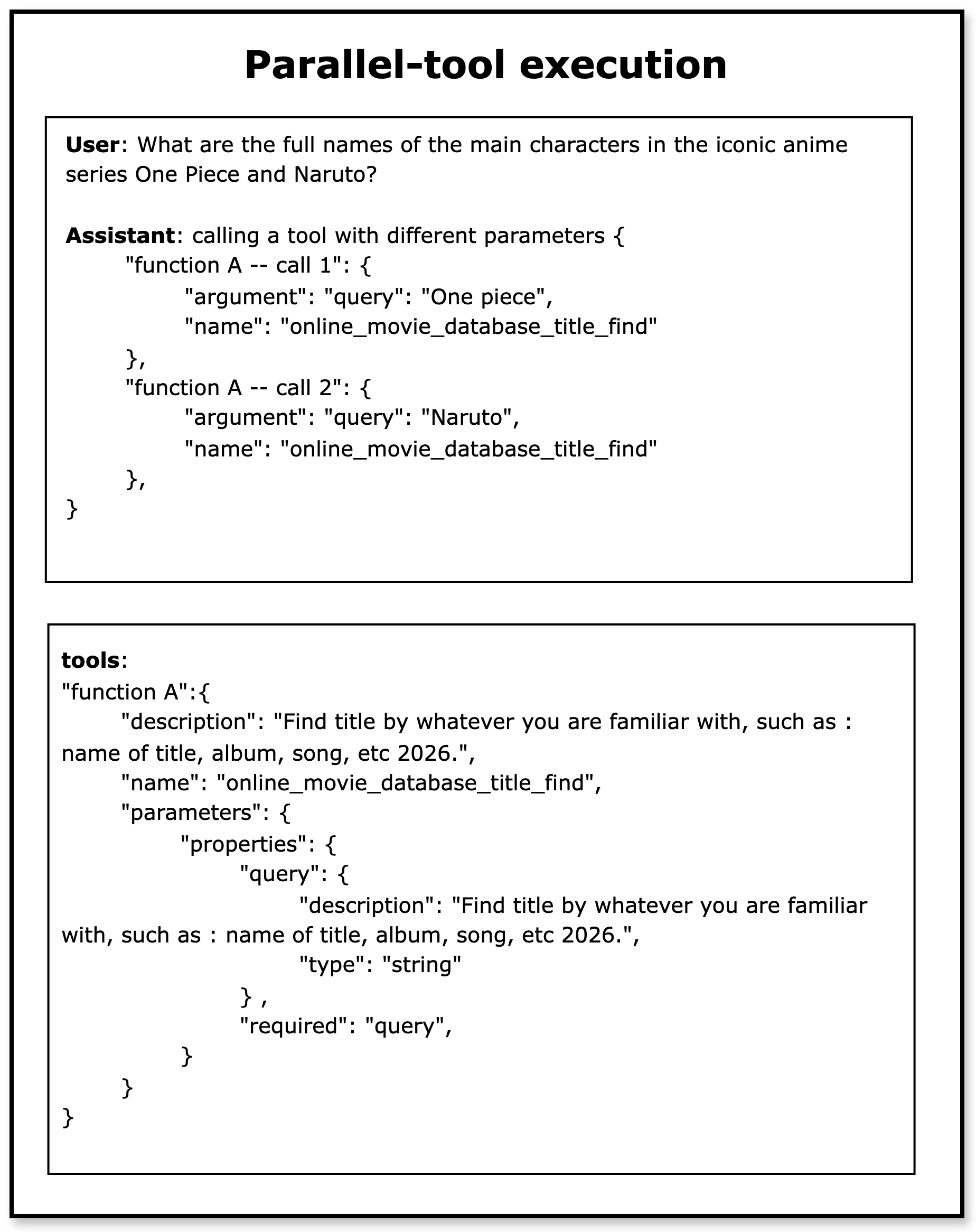}
    \caption{Example of parallel tool-use}
    \label{fig:parallel-tool-use-samples}
\end{figure}

\begin{figure}
    \centering
    \includegraphics[width=.85\textwidth]{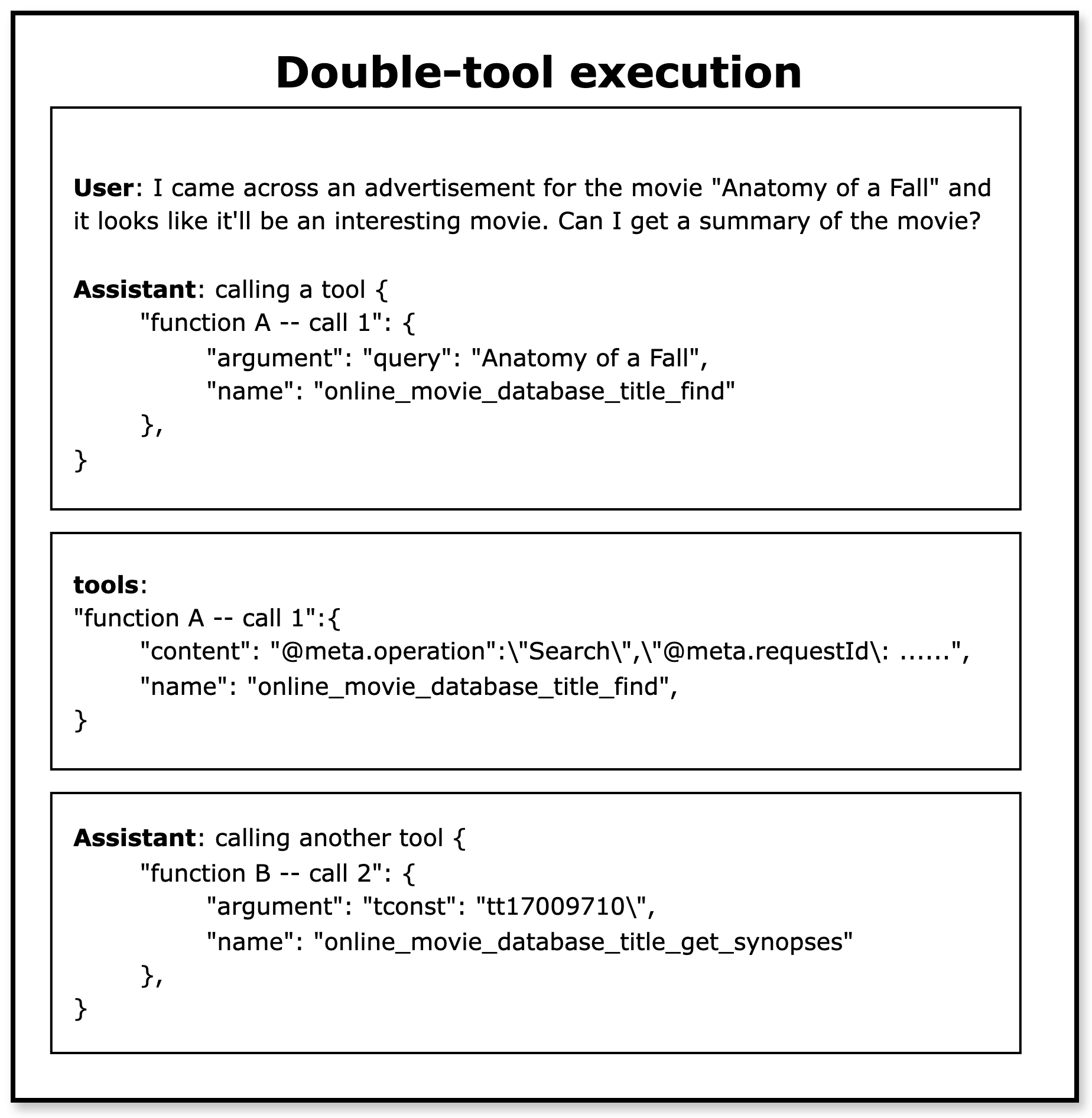}
    \caption{Example of multi-tool use}
    \label{fig:multi-tool-use-samples}
\end{figure}

For tool use the concrete prompts are generated by the model to transform the tool definitions as well as the predicted tool calls and tool results into the concrete prompt. Therefore, we don't report concrete prompt here. Samples from the dataset are show in Figs.~\ref{fig:single-tool-use-samples}, \ref{fig:parallel-tool-use-samples}, and \ref{fig:multi-tool-use-samples}.

\subsection{DAG}

\begin{lstlisting}[
caption={DAG Reasoning Prompt}, label={lst:dag_reasoning_prompt}]
Choose the appropriate plan with its associated set of tools to accomplish a specific task.

Each tool's functionality is using Open API schema notation.

"[A B]" indicates that tools A and B can operate concurrently without any interdependencies.
"A -> B" indicates that tool A must be executed prior to tool B, as B's operation is contingent upon the output from A.
"B -> A" indicates that tool B must be executed prior to tool A, as A's operation is contingent upon the output from B.
"[A B] -> C" indicates that tools A and B will be executed in parallel initially, followed by tool C, which requires the outputs from both A and B.

Below is a compilation of all available tools, described in JSON format:

{tool_descriptions}

Here are all possible plans:

{possible_plans}

Task Input:

{task_input}


Feel free to elaborate on your thought process and reasoning.

Please strictly end the reply with "Chosen Plan: N", where "N" must be an integer number of the selected plan.
Do not change the answer format to other words and MUST include Chosen Plan:
\end{lstlisting}

\begin{lstlisting}[
caption={DAG Planning Prompt}, label={lst:dag_planning_prompt}]
Choose the appropriate plan with its associated set of tools to accomplish a specific task.

Each tool's functionality is using Open API schema notation.

"[A B]" indicates that tools A and B can operate concurrently without any interdependencies.
"A -> B" indicates that tool A must be executed prior to tool B, as B's operation is contingent upon the output from A.
"B -> A" indicates that tool B must be executed prior to tool A, as A's operation is contingent upon the output from B.
"[A B] -> C" indicates that tools A and B will be executed in parallel initially, followed by tool C, which requires the outputs from both A and B.

Below is a compilation of all available tools, described in JSON format:

{tool_descriptions}

Here are all possible plans:

{possible_plans}

Task Input:

{task_input}


Be concise with a response in the format "Chosen Plan: N", where "N" represents the number of the selected plan.
\end{lstlisting}

\subsection{Data Science and Machine Learning}
\begin{lstlisting}[
caption={DS \& ML Code Generation Prompt}, label={lst:code_kaggle_user_prompt}]
You are a helpful code assistant that will help users analyze data and answering multiple choice questions based on the data analysis.
You will be provided with a Jupyter notebook file (.ipynb) containing code cells and markdown cells.
A list of data files used in the notebook will also be provided.
The notebook would contain a code interpreter style conversation between a user and an assistant.
User would be asking questions, follow up with additional requests, and providing clarification for requests.
Assistant would be analyzing data with code interpreter, providing insights into data based on analysis results and diagrams, and responding to user inquiries.

Your task is to address the user's request in the last markdown cell of the notebook AND answer the multiple choice questions based on the data analysis.
You will first append an appropriate code cell to the notebook to address the user's request.
You will then be provided with the execution results of the code cell. Based on the results, you will answer the multiple choice questions.

For example, given

Data Files:
['../input/pokemon/pokemon.csv']

Notebook:
{"cells": [{"cell_type": "markdown", "id": "188d8f78", "metadata": {}, "source": ["**user**\n", "\n", "Can we start by loading the Pok\u00e9mon dataset and displaying the first few rows?"]}, {"cell_type": "markdown", "id": "41316f22", "metadata": {}, "source": ["**assistant**\n", "\n", "Absolutely, let's load the Pok\u00e9mon dataset using pandas and display the first three rows."]}, {"cell_type": "code", "execution_count": 1, "id": "d07755b0", "metadata": {"execution": {"iopub.execute_input": "2024-04-12T14:11:29.209092Z", "iopub.status.busy": "2024-04-12T14:11:29.207876Z", "iopub.status.idle": "2024-04-12T14:11:30.934431Z", "shell.execute_reply": "2024-04-12T14:11:30.932965Z"}}, "source": ["\n", "import pandas as pd\n", "\n", "# Load the Pok\u00e9mon dataset\n", "pokemon = pd.read_csv(\"../input/pokemon/pokemon.csv\")\n", "pokemon.head(3)\n", "    "]}], "metadata": {"language_info": {"codemirror_mode": {"name": "ipython", "version": 3}, "file_extension": ".py", "mimetype": "text/x-python", "name": "python", "nbconvert_exporter": "python", "pygments_lexer": "ipython3", "version": "3.11.8"}}, "nbformat": 4, "nbformat_minor": 5}

User Request:
I'm interested in visualizing the frequency of Pok\u00e9mon by their primary type. Can you help with that?

You will respond with markdown cell and code cell to address the user's request:
[{"cell_type": "markdown", "metadata": {}, "source": ["**assistant**\n", "\n", "Sure thing! Let's create a bar chart to visualize the frequency of Pok\u00e9mon by their primary type."]}, {"cell_type": "code", "metadata": {}, "source": ["\n", "import matplotlib.pyplot as plt\n", "\n", "# Frequency of Pok\u00e9mon by their primary type\n", "pokemon['type1'].value_counts().plot.bar()\n", "plt.title('Frequency of Pok\u00e9mon by Primary Type')\n", "plt.xlabel('Primary Type')\n", "plt.ylabel('Frequency')\n", "plt.show()\n", "    "]}]

Then, you will be provided with the execution results of the code cell and multiple choice questions based on the results:

Execution Output:  # the images would be attached in the same order as the image placeholders
[{"data": {"image/png": "<image_1>"}, "execution_count": 1, "metadata": {}, "output_type": "display_data"}]

Questions:
[{"question": "Which primary type of Pok\u00e9mon has the highest frequency according to the chart?", "choices": {"A": "Rock", "B": "Water", "C": "Fire", "D": "Grass"}}, {"question": "What type of Pok\u00e9mon is least frequent as per the visualization?", "choices": {"A": "Ghost", "B": "Fairy", "C": "Ice", "D": "Dragon"}}, {"question": "According to the bar chart, which primary type has more Pok\u00e9mon than Fairy but fewer than Fire?", "choices": {"A": "Electric", "B": "Grass", "C": "Ice", "D": "Dragon"}}, {"question": "What primary type appears more frequently than 'Rock' but less than 'Grass' according to the visualization?", "choices": {"A": "Ghost", "B": "Poison", "C": "Flying", "D": "Bug"}}]

You will then provide the answers to the multiple choice questions:
["B", "B", "C", "B"]

You MUST exactly follow nbformat structure when generating markdown and code cells.
You are only allowed to use the following math and visualization packages: numpy, pandas, scipy, scikit-learn, sympy, statsmodels, matplotlib, seaborn, plotly, wordcloud. Python standard library is also allowed. Make sure to import the necessary packages in the code cell.
You MUST answer in a list of uppercase single letter strings for multiple choice questions.

Now, given the following notebook and user request, respond with a list containing a markdown cell and a code cell to address the user's request:

Data Files:
{data_files}

Notebook:
{notebook}

User Request:
{user_request}



You MUST exactly follow nbformat structure when generating markdown and code cells.
Respond ONLY with the cells in json format. DO NOT say anything else or put it in a ```json``` code block.
You are only allowed to use the following math and visualization packages: numpy, pandas, scipy, scikit-learn, sympy, statsmodels, matplotlib, seaborn, plotly, wordcloud. Python standard libraries are also allowed. Make sure to import the necessary packages in the code cell.
\end{lstlisting}

\begin{lstlisting}[
caption={DS \& ML QA Prompt}, label={lst:code_kaggle_qa_prompt}]
Given the following execution output and multiple choice questions, respond with the answers to the questions:

Execution Output:
{execution_output}

Questions:
{questions}

You MUST answer in a list of uppercase single letter strings for multiple choice questions.
\end{lstlisting}

\begin{lstlisting}[
caption={DS \& ML Execution Feedback Prompt}, label={lst:code_kaggle_execution_feedback_prompt}]
Your code execution failed. Please check the error message below and try again:

{error_message}

You MUST exactly follow nbformat structure when generating markdown and code cells.
Respond ONLY with the cells in json format. DO NOT say anything else or put it in a ```json``` code block.
You are only allowed to use the following math and visualization packages: numpy, pandas, scipy, scikit-learn, sympy, statsmodels, matplotlib, seaborn, plotly, wordcloud. Python standard libraries are also allowed. Make sure to import the necessary packages in the code cell.
\end{lstlisting}

\subsection{Contest-Level Coding}

\begin{lstlisting}[
caption={CodeContests E2E Standard Prompt}, label={lst:code_contests_regular_prompt}]
You are an AI with advanced code generation capabilities.

When you encounter a specific problem (labeled #PROBLEM#), your goal is to produce a valid python code that correctly solves the problem.
Make sure to fully address the problem goals following the rules and constraints.
The code should be robust and general. It should output correct answers under any valid inputs, not only just the example inputs given in the problem description.

#PROBLEM#:
{description}

The code output must follow this structure:
```
def f1(...):
    ...
    return ...

def f2(...):
    ...
    return ...
...

if __name__ == "__main__":
    ...
```
The code should read the input using the 'input()' method. Make sure to properly parse the input, according to the problem description.
The output should be printed without additional words using the 'print()' method.

#ANSWER#:
```python
\end{lstlisting}

\begin{lstlisting}[
caption={CodeContests Problem Parsing Prompt}, label={lst:code_contests_understand_prompt}]
You are an AI with advanced comprehension capabilities, akin to that of humans. This enables you to understand complex instructions, discern the user's intent, and apply logical reasoning across a variety of scenarios effectively.

When you encounter a specific problem (labeled #PROBLEM#) along with a relevant example input (referred to as #TEST CASE#), your goal is to deduce the correct answer (#ANSWER#).

Please adhere to the following format:
First, express your #THOUGHTS# to reflect your understanding of the problem and test case. Keep your #THOUGHTS# concise, not exceeding 5 sentences.
Then, provide your #ANSWER# to the problem.
For example, your output can be:
#THROUGHTS#:
The problem is about ... The test cases are ...
#ANSWER#:
1 2
3 4

Here is the problem.

#PROBLEM#:
{description}

#TEST CASE#:
{inputs}

Now, please provide the #THOUGHTS# and #ANSWER# sections. Do not include additional explanations or reasoning processes.
#THROUGHTS#:
\end{lstlisting}

\begin{lstlisting}[
caption={CodeContests Planner-shift Prompt}, label={lst:code_contests_plan_prompt}]
You are an AI with advanced code understanding and planning capabilities.

When you encounter a specific problem (labeled #PROBLEM#), your goal is to devise a structured and executable plan to solve the problem.
Your plan should clearly outline the logical steps needed to address the problem, breaking down complex processes into manageable actions.
It's crucial that each step is straightforward enough to be easily translated into executable code or functions by a Language Model (LLM).
Keep the plan simple and focused, avoiding unnecessary complexity to ensure ease of implementation.

#PROBLEM#:
{description}

#GENERATED PLAN#:
\end{lstlisting}

\begin{lstlisting}[
caption={CodeContests Solver-shift Prompt}, label={lst:code_contests_solve_prompt}]
You are an AI with advanced code generation and instruction following capabilities.

When you encounter a specific problem (labeled #PROBLEM#) and a solving plan (labeled #PLAN#), your goal is to produce a valid python code that correctly solves the problem.
Make sure to fully address the problem goals following the rules and constraints. Refer to the plan to generate the code.
The code should be robust and general. It should output correct answers under any valid inputs, not only just the example inputs given in the problem description.

#PROBLEM#:
{description}

#PLAN#:
{plan}

guidelines:
- Generate only code, without any additional explanations or comments.
- Make sure to include all the necessary module imports, properly initialize the variables, and address the problem constraints.
- The code needs to be self-contained, and executable as-is.

The code output must follow this structure:
```
def f1(...):
    ...
    return ...

def f2(...):
    ...
    return ...
...

if __name__ == "__main__":
    ...
```
The code should read the input using the 'input()' method. Make sure to properly parse the input, according to the problem description.
The output should be printed without additional words using the 'print()' method.

answer:
```python
\end{lstlisting}

\subsection{Mathematics}

\begin{lstlisting}[
caption={Math Generation Prompt}, label={lst:math_generation_prompt}]
You are an AI with advanced comprehension capabilities, akin to that of humans. This enables you to understand complex instructions, discern the user's intent, and apply logical reasoning across a variety of scenarios effectively.

When you encounter a specific problem (labeled #PROBLEM#), your goal is to deduce the correct answer (#ANSWER#).

Please adhere to the following format:
First, express your #THOUGHTS# to reflect your understanding of the problem. Keep your #THOUGHTS# concise, not exceeding 5 sentences.
Then, provide your #ANSWER# to the problem.
For example, your output can be:
#THROUGHTS#:
Let's think step by step ...
#ANSWER#:
12

Here is the problem.

#PROBLEM#:
{description}

Now, please provide the #THOUGHTS# and #ANSWER# sections. Do not include additional explanations or reasoning processes under #ANSWER# section.
#THROUGHTS#: Let's think step by step:

#ANSWER#:
\end{lstlisting}

\begin{lstlisting}[
caption={Math Evaluation Prompt}, label={lst:math_evaluation_prompt}]
As a math expert, you will be provided with three items: a #Question#, an #Answer#, and the #Ground Truth#.
Your task is to determine whether the #Answer# matches the #Ground Truth#.
You need to considering mathmetical theorems when verifing the answer. For example, '$(c + 9)(c - 4)$' and '(c - 4)*(c + 9)'should be
the same expressions based on factorization theorems.
If they align, respond with 'correct'. If they do not, respond with 'wrong'.
Ensure that your #Response# is the final line and consists solely of the word 'correct' or 'wrong', without any additional commentary or explanation.

#Question#:
{question}

#Answer#:
{answer}

#Ground Truth#:
{ground_truth}

#Response#:
\end{lstlisting}

\begin{lstlisting}[
caption={Math Planning Prompt}, label={lst:math_planning_prompt}]
You are a math planner.
Given a problem description, your goal is to devise a structured and executable plan to solve the problem.
Your plan should clearly outline the logical steps needed to address the problem, breaking down complex processes into manageable actions.
It's crucial that each step is straightforward enough to be easily translated into executable calculation or functions by a Language Model (LLM).
Keep the plan simple and focused, avoiding unnecessary complexity to ensure ease of implementation.
Problem description:
=============
{description}
=============

Generated plan:
\end{lstlisting}
\clearpage
\section{More evaluation results}
\label{sec:appendix_eval_results}

For domain-centric evaluation, we chose tasks that rely solely on the evaluation model to solve them end-to-end and adhere to the standard setup used in the field. This ensures that the domain-centric results accurately reflect the most standardized evaluation outcomes, making it easier to compare performance across different benchmarks. Specifically, we use the evalution results of \texttt{E2E standard} task as domain evaluation results for \textcolor{teal}{Contest-level coding}, \textcolor{teal}{DS$\&$ML}, \textcolor{teal}{Math} and \textcolor{teal}{DAG-QA}. For \textcolor{teal}{Tool-use}, we report the weighted average of single-, parallel-, and multi-turn multi-tool-use tasks. The results are shown in Table~\ref{tab:domain_centric_eval_results}.

\begin{table}[h!]
    \tiny
    \centering
    \caption{Evaluation resutls on math tasks. stand.: standard E2E, underst.: comprehend+, solv.-shift: solver-shift, plan.-shift: planner-shift.}
    \begin{tabular}{lcccc|lcccc}
    \toprule
         \rowcolor{tbgray} Commercial Models & stand. & compre. & solv.-shift & plan.-shift & OS Models  & stand. & compre. & solv.-shift & plan.-shift \\ \hline
         GPT-4o & 53.4 & 51.48 & 61.2 & 45.2 & Mixtral-8x22B & 50 & 42.46 & 60.02 & 40.20  \\
         GPT-4-turbo & 38.57 & 36.72 & 50.32 & 46.60 & Mixtral-8x7B & 21.70 & 27.12 & 43.46 & 28.00 \\
         GPT-3.5-turbo &25.00 & 20.41 & 45.80 & 32.80 & Mistral-7B & 9.55 & 25.00 & 27.12 & 15.63 \\
         Gemini-1.5-pro & 39.70 & 35.21 & 56.60 & 25.81  & Phi-3 & 21.00 & 21.00 & 34.23 & 31.30  \\
         Gemini-1.0-pro & 39.40 & 33.13 & 66.67 & 41.67 & openfunctions-v2 & 15.75 & 13.31 & 25.00 & 23.49 \\
         Claude-3-opus & 37.40 & 36.09 & 56.00 & 36.80 & Hermes-2-Pro &12.78 & 12.95 & 39.62 & 25.73 \\
         Claude-3-sonnet & 26.80 & 25.44 & 42.90 & 32.30  & Llama-2-70b & 8.43 & 8.29 & 32.10 & 11.40 \\
         Claude-3-haiku & 36.60 & 27.51 & 44.50 & 49.65  & Llama-2-13b  & 4.10 & 3.85 & 19.84 & 3.85 \\ 
         Command-R & 8.21 & 10.00 & 33.33 & 23.70 & Llama-2-7b & 3.92 & 2.23 & 24.69 & 25.73 \\ 
    \bottomrule
    \end{tabular}
    \label{tab:math_planner_solver_shift_eval}
\end{table}

Detailed tool use results are reported in Table \ref{tab:tool-use-details}.

\begin{table}[h!]
    \centering
    % \small 
    \caption{Detailed results for the tool use datasets. Owing to incorrect filtering, the distractor datasets contain two additional samples. They will be removed in follow-up versions.}
    \resizebox{\textwidth}{!}{
    \begin{tabular}{l|lccccccc}
    \toprule
    Dataset                     & Single-step
                                & Parallel 
                                & \begin{tabular}[c]{@{}c@{}} Multi-step  \\ Double         \end{tabular}
                                & \begin{tabular}[c]{@{}c@{}} Multi-step  \\ Triple         \end{tabular}
                                & \begin{tabular}[c]{@{}c@{}} Single-step \\ w. Distractors \end{tabular}
                                & \begin{tabular}[c]{@{}c@{}} Parallel    \\ w. Distractors \end{tabular} 
                                & \begin{tabular}[c]{@{}c@{}} Multi-step  \\ Double \\ w. Distractors \end{tabular}
                                & \begin{tabular}[c]{@{}c@{}} Multi-step  \\ Triple \\ w. Distractors \end{tabular} \\
    \# Samples                  & 184   & 225   & 298  & 258    & 185   & 225   & 300   & 258   \\
    \midrule
    Command R+                  & 0.739 & 0.662 & -     & -     & 0.708 & 0.644 & -     & -     \\
    Command R                   & 0.717 & 0.627 & -     & -     & 0.654 & 0.609 & -     & -     \\
    Claude 3 Haiku              & 0.837 & 0.053 & 0.513 & 0.415 & 0.768 & 0.040 & 0.433 & 0.357 \\
    Claude 3 Opus               & 0.886 & 0.436 & 0.708 & 0.504 & 0.854 & 0.387 & 0.663 & 0.535 \\
    Claude 3 Sonnet             & 0.826 & 0.302 & 0.581 & 0.419 & 0.811 & 0.169 & 0.463 & 0.380 \\
    Gemini-1.0-pro              & 0.750 & 0.004 & 0.252 & 0.155 & 0.708 & 0.004 & 0.197 & 0.151 \\
    Gemini-1.5-flash            & 0.783 & 0.191 & 0.534 & 0.407 & 0.784 & 0.187 & 0.467 & 0.391 \\
    Gemini-1.5-pro              & 0.772 & 0.440 & 0.530 & 0.415 & 0.735 & 0.471 & 0.473 & 0.376 \\
    Openfunctions-v2            & 0.875 & 0.422 & -     & -     & 0.832 & 0.373 & -     & -     \\
    GPT-3.5-Turbo               & 0.918 & 0.667 & 0.634 & 0.523 & 0.881 & 0.609 & 0.620 & 0.457 \\
    GPT-4-Turbo                 & 0.918 & 0.853 & 0.715 & 0.585 & 0.881 & 0.818 & 0.653 & 0.531 \\
    GPT-4o-2024-05-13           & 0.913 & 0.822 & 0.607 & 0.523 & 0.881 & 0.809 & 0.543 & 0.465 \\
    Hermes-2-Pro-Mistral-7B     & 0.750 & 0.556 & 0.245 & 0.174 & 0.746 & 0.498 & 0.157 & 0.128 \\
    \bottomrule
    \end{tabular}
    }
    
    % \caption{Detailed results for the tool use datasets. Owing to incorrect filtering, the distractor datasets contain two additional samples. They will be removed in follow-up versions.}
    \label{tab:tool-use-details}
\end{table}

% For some samples the arguments to a tool can be highly ambiguous. Therefore, we have introduced detailed argument matching rules based on regular expression. They allow for a weaker comparison than exact string matching. An example might be "I'm thinking of relocating to Boston from France due to some personal reasons. Can you get me the name and address of a 3-bedroom apartment". In this case the model needs to call a tool for searching apartments and use Boston as the location. Some models tend to generate "Boston, MA" and a pure string based match would report a bad prediction here. That is avoided by using a regular expression. I.e. if the model calls the correct tool with a an argument containing Boston our evaluation would still report that sample as a successful prediction.

For Command R, Command R+, and Hermes-2-Pro-Mistral-7B we used the tool use prompts as reported in the official model repositories \footnote{\url{https://huggingface.co/CohereForAI/c4ai-command-r-v01}} \footnote{\url{https://huggingface.co/CohereForAI/c4ai-command-r-plus}} \footnote{\url{https://huggingface.co/NousResearch/Hermes-2-Pro-Mistral-7B}}. These prompts for Command R and Command R+ do not support multi-step tool calls, so we could not evaluate Command R and Command R+ on the double and triple datasets.

The Claude models as well as the Gemini and GPT models support tool use through dedicated APIs. These APIs have been used for the evaluation here.

\textbf{Detailed Reulst on Data Science and Machine Learning}

\begin{table*}[!tbp]
    \caption{Detailed evaluation results on Data Science and Machine Learning. The bolded and underlined numbers indicate the 1st and 2nd highest performances in each category.}
\small
\centering
\resizebox{1.0\textwidth}{!}{
\begin{tabular}{lccccccc}
\toprule
 \rowcolor{tbgray} Model & E2E & GPT4 QA & Oracle Execution & Retry \\
\midrule
GPT-4o~\cite{gpt-4o} & \textbf{66.52} & \textbf{64.27} & \textbf{88.60} & \textbf{70.79} \\
GPT-4-Turbo~\cite{gpt4} & \underline{63.90} & \underline{63.90} & \underline{85.67} & \underline{68.05} \\
GPT-3.5-Turbo & 35.79 & 58.78 & 52.13 & 37.44 \\
Gemini-1.5-pro~\cite{reid2024gemini} & 54.02 & 55.55 & 83.23 & 58.90 \\
Gemini-1.0-pro~\cite{team2023gemini} & 27.87 & 45.12 & 52.93 & 28.66 \\ 
Claude3 Opus~\cite{claude3} & 59.45 & 61.65 & 83.48 & 65.12 \\
Claude3 Sonnet~\cite{claude3} & 52.44 & 56.71 & 75.37 & 54.33 \\
Claude3 Haiku~\cite{claude3} & 33.84 & 48.48 & 59.39 & 36.04 \\
\midrule
Mixtral-8x22B-v0.1~\cite{mixtral} & 31.10 & 45.43 & 58.05 & 31.65 \\
Mixtral-8x7B-v0.1~\cite{mixtral} & 13.48 & 26.46 & 48.90 & 13.90 \\
Mistral-7B-v0.2~\cite{jiang2023mistral} & 2.80 & 19.94 & 12.40 & 3.17 \\
Phi-3-mini4K-instruct~\cite{phi3} & 0.84 & 4.34 & 0.00 & 3.35 \\
Openfunctions-v2~\cite{gorilla-openfunctions-v2} & 3.84 & 17.07 & 12.74 & 4.02 \\
Hermes-2-Pro-Mistral-7B~\cite{Hermes-2-Pro-Mistral-7B} & 3.29 & 17.93 & 23.02 & 3.35 \\
Command R~\cite{command-r} & 0.00 & 0.00 & 48.88 & 0.00 \\
LLama2-70B~\cite{touvron2023llama2} & 0.00 & 19.51 & 0.00 & 0.00 \\
Llama2-13B~\cite{touvron2023llama2} & 0.00 & 0.00 & 0.00 & 0.00 \\
Llama2-7B~\cite{touvron2023llama2} & 0.00 & 0.00 & 0.00 & 0.00 \\
\bottomrule
\end{tabular}}%
\label{tab:dsml_result}
\end{table*}

Table~\ref{tab:dsml_result} presents the evaluation results on Data Science and Machine Learning tasks, assessing E2E code generation and QA, code generation with GPT-4 QA, QA from oracle code execution, and DS \& ML self-correction.

GPT-4o achieves the highest accuracy across all settings: E2E (66.52\%), Execution Only + GPT-4 QA (64.27\%), QA from Oracle Execution (88.60\%), and E2E + Online Retry (70.79\%). GPT-4-Turbo follows closely with 63.90\%, 63.90\%, 85.67\%, and 68.05\% respectively. These results highlight their robust code generation and QA capabilities.

Gemini-1.5-pro and Claude3 Opus show strong performance in QA from Oracle Execution with 83.23\% and 83.48\%, indicating good comprehension and QA skills, though their code generation is less effective compared to GPT-4 models. Mixtral-8x22B-v0.1 has the overall best performance among the evaluated open-source models.

% Models like Openfunctions-v2, Phi-3-mini4K-instruct, and Llama2 variants show low accuracy across tasks, struggling particularly with QA. For instance, LLama2-70B scores 0.00\% in E2E.

% Overall, while GPT-4o and GPT-4-Turbo lead in performance, the variability across tasks and models highlights areas for improvement in integrating code generation with QA.

\textbf{Detailed Results on Contest-level Coding.}

\begin{table*}[!tbp] % [!h]  % tbp
    \caption{Detailed evaluation results on Contest-level Coding. The bolded and underlined numbers indicate the 1st and 2nd highest performances in each category.}
% \small
\centering
\resizebox{1.0\textwidth}{!}{
\begin{tabular}{lccccccc}
\toprule
 \rowcolor{tbgray} Model &   E2E Standard        & Problem Parsing & PlannerShift &     SolverShift &    Retry   \\
\midrule
GPT-4o~\cite{gpt-4o} & \textbf{31.80} & \textbf{49.04} & \textbf{24.13} & \textbf{26.44} & \textbf{9.55}  \\
GPT-4-Turbo~\cite{gpt4} & \underline{25.67} & \underline{38.31} & 18.01 & \underline{19.54} & \underline{6.7}  \\
GPT-3.5-Turbo & 10.34 & 30.65 & 17.60 & 11.49 & 0.85  \\
Gemini-1.5-pro~\cite{reid2024gemini} & 10.34 & 34.87 & 16.09 & 11.88 & 4.27 \\
Gemini-1.0-pro~\cite{team2023gemini} & 7.66 & 27.20 & \underline{19.92} & 9.58 & 0 \\ 
Claude3 Opus~\cite{claude3}      & 15.33 & 39.08 & 21.07 & 12.26 & 4.98  \\
Claude3 Sonnet~\cite{claude3}    & 10.34 & 34.10 & 16.48 & 10.73 & 1.71 \\
Claude3 Haiku~\cite{claude3}     & 8.81 & 32.57 & 22.99 & 10.73 & 0.84  \\
\midrule
Mixtral-8x22B-v0.1~\cite{mixtral}    &  \textbf{9.20} & \textbf{32.18} & 15.32 & \textbf{11.49} & \underline{0.84}  \\
Mixtral-8x7B-v0.1~\cite{mixtral}      & 1.92 & 22.22 & 20.31 & 4.21 & \textbf{1.17} \\
Mistral-7B-v0.2~\cite{jiang2023mistral}   & 0.38 & 8.05 & 21.84 & 4.21 & 0.38 \\
Phi-3-mini4K-instruct~\cite{phi3} & 2.30 & 18.01 & \textbf{24.52} & 3.07 & 0.39  \\
Openfunctions-v2~\cite{gorilla-openfunctions-v2}  & 8.05 & 3.83 & 19.16 & \underline{9.96} & 0.83 \\
Hermes-2-Pro-Mistral-7B~\cite{Hermes-2-Pro-Mistral-7B} & 0.77 & 17.62 & \underline{23.37} & 3.83 & 0.39  \\
Command R~\cite{command-r}      & 4.21 & 6.90 & 18.00 & 7.28 & 0.80  \\
LLama2-70B~\cite{touvron2023llama2} & 0  & 9.20 & 22.61 & 0 & 0.77  \\
Llama2-13B~\cite{touvron2023llama2} & 0  & 1.53 & 21.07 & 2.30 & 0  \\
Llama2-7B~\cite{touvron2023llama2} & 0 & 5.75 & 20.31 & 2.30 & 0 \\
\bottomrule
\end{tabular}}%
% \vspace{-2.5em}
\label{tab:codecontests_result}
\end{table*}

Table~\ref{tab:codecontests_result} shows detailed results in the Contest-level Coding domain. As a challenging dataset, we find that the overall E2E Standard scores are relatively low across models, while GPT-4o and GPT-4-Turbo achieve much better results than remaining models by a large margin. GPT-4o has leading performance over all types of tasks, especially Problem Parsing and Retry, showing its remarkable capability of understanding and self-correct. \\
With a separation of capabilities, we can better analyze the differences and gaps among existing models. 
For example, we observe significant performance variation across different models on the Problem Parsing task, suggesting the intellectual challenge of understanding the contest problems due to the complicated nature of the problem description. On the other hand, we emphasize that the design of E2E Standard, SolverShift and Retry tasks require the model to generate code, while Problem Parsing and PlannerShift eliminate the actual code generation step and mainly focus on understanding/reasoning/planning capabilities. 
Such a fine-grained model evaluation can reflect more detailed differences between models. We can see that although some open-sourced models fail to solve most E2E Standard problems, they still possess good understanding and planning capabilities on the complicated contest-level problems (e.g. Phi-3-mini4K-instruct, Hermes-2-Pro-Mistral-7B). \\
By comparing the E2E Standard and Planner-shift or Solver-shift, we can see that most models have boosted performance with a separate strong solver or planner. However, the pass rate of GPT-4o and GPT-4-Turbo drops on Planner-shift and Solver-shift compared to their E2E Standard performance. This may be caused by the selection of GPT-4-0125 as the fixed solver/planner, which itself has 22.61\% score on E2E Standard, stronger than most models while weaker than GPT-4o and GPT-4-Turbo. \\
Moreover, we find that for Contest-level Coding, having a plain step-by-step plan does not bring as significant improvement as in the math domain, as such plans may not be sufficient to solve the algorithmic challenges in the domain. A more powerful planning and reasoning structure may be adopted to better solve the Contest-level problems, as shown in \cite{alphacodium:ridnik2024code}. But the focus of this paper is to identify and reveal the capabilities of existing LLMs, which advancing the capabilities remains as a research problem for future investigation.

\newpage
\section{Data examples}
\label{sec:appendix_data_examples}

% \subsection{Tool use self-correction}

\begin{figure} [h!]
    \centering
    \includegraphics[width=\textwidth]{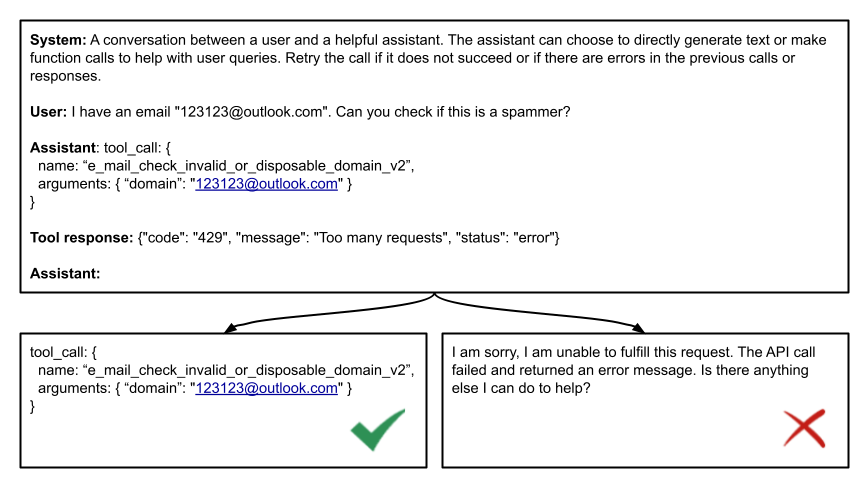}
    \caption{Retry example where the assistant needs to recognize a temporary error and retry the call.}
    \label{fig:tool-use-retry-temp}
\end{figure}

\begin{figure} [h!]
    \includegraphics[width=\textwidth]{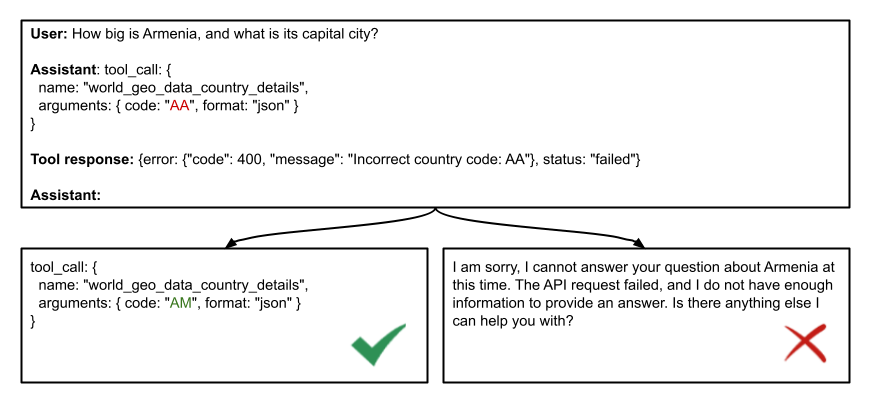}
    \caption{Self-correction example where the assistant needs to correct an erroneous call given explicit feedback from the tool.}
    \label{fig:tool-use-retry-error}
\end{figure}

\begin{figure}
    \includegraphics[width=\textwidth]{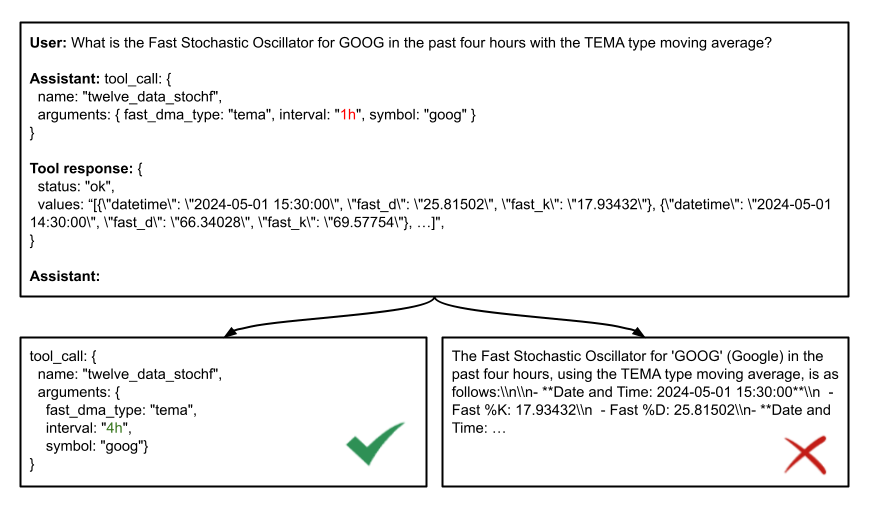}
    \caption{Self-correction example where the assistant needs to recognize that the prior call, despite succeeding, did not match the user's request, and re-issues the corrected call.}
    \label{fig:tool-use-retry-silent}
\end{figure}

% \begin{figure}[h]
%     \centering
%     \includegraphics[width=.8\textwidth]{figures/tool-use-examples.png}
%     \caption{\small {Examples from the tool use dataset.}}
%     \label{fig:tool-use-examples}
% \end{figure}

\begin{figure}
    \centering
    \includegraphics[width=.8\textwidth]{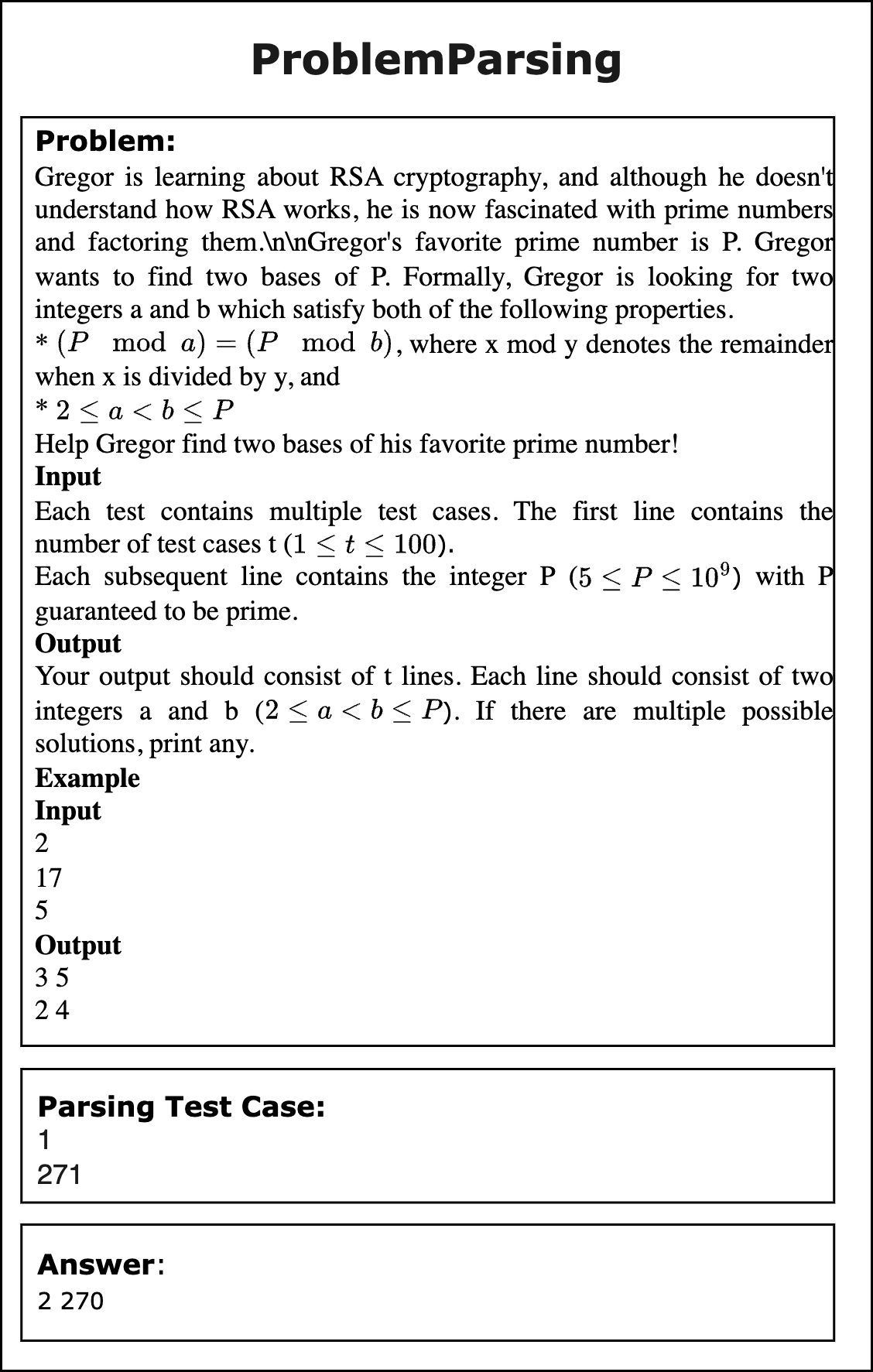}
    \caption{\small {An example of CodeContests ``ProblemParsing'' task to measure the agent's understanding capability.}}
    \label{fig:codecontests-understand-problem}
\end{figure}
\clearpage

\end{document}